# Sequential Diagnosis by Abstraction

**Sajjad Siddiqi**
*National University of Sciences and Technology*
*(NUST) Islamabad, Pakistan*                     SAJJAD.AHMED@SEECS.EDU.PK

**Jinbo Huang**
*NICTA and Australian National University*
*Canberra, Australia*                            JINBO.HUANG@NICTA.COM.AU

## Abstract

When a system behaves abnormally, *sequential diagnosis* takes a sequence of measurements of the system until the faults causing the abnormality are identified, and the goal is to reduce the *diagnostic cost*, defined here as the number of measurements. To propose measurement points, previous work employs a heuristic based on reducing the entropy over a computed set of *diagnoses*. This approach generally has good performance in terms of diagnostic cost, but can fail to diagnose large systems when the set of diagnoses is too large. Focusing on a smaller set of probable diagnoses scales the approach but generally leads to increased average diagnostic costs. In this paper, we propose a new diagnostic framework employing four new techniques, which scales to much larger systems with good performance in terms of diagnostic cost. First, we propose a new heuristic for measurement point selection that can be computed efficiently, without requiring the set of diagnoses, once the system is modeled as a Bayesian network and compiled into a logical form known as d-DNNF. Second, we extend *hierarchical diagnosis*, a technique based on system abstraction from our previous work, to handle probabilities so that it can be applied to sequential diagnosis to allow larger systems to be diagnosed. Third, for the largest systems where even hierarchical diagnosis fails, we propose a novel method that converts the system into one that has a smaller abstraction and whose diagnoses form a superset of those of the original system; the new system can then be diagnosed and the result mapped back to the original system. Finally, we propose a novel cost estimation function which can be used to choose an abstraction of the system that is more likely to provide optimal average cost. Experiments with ISCAS-85 benchmark circuits indicate that our approach scales to all circuits in the suite except one that has a flat structure not susceptible to useful abstraction.

## 1. Introduction

When a system behaves abnormally, the task of *diagnosis* is to identify the reasons for the abnormality. For example, in the combinational circuit in Figure 1, given the inputs $P \wedge Q \wedge \neg R$, the output $V$ should be 0, but is actually 1 due to the faults at gates $J$ and $B$. Given a system comprising a set of components, and a knowledge base modeling the behavior of the system, along with the (abnormal) observed values of some system variables, a (consistency-based) *diagnosis* is a set of components whose failure (assuming the other components to be healthy) together with the observation is logically consistent with the system model. In our example, $\{V\}, \{K\}, \{A\}$, and $\{J, B\}$ are some of the diagnoses given the observation. In general, the number of diagnoses can be exponential in the number of system components, and only one of them will correspond to the set of actual faults.

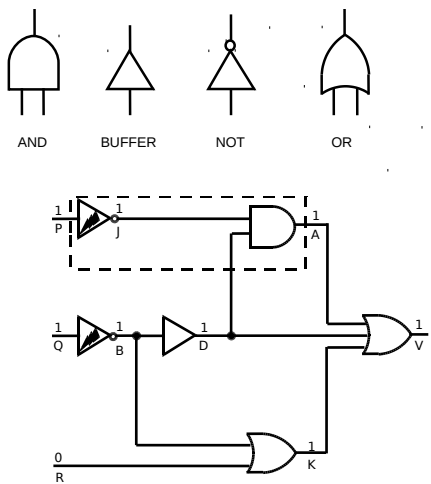

Figure 1: A faulty circuit.

In this paper, therefore, we consider the problem of *sequential diagnosis* (de Kleer & Williams, 1987), where a sequence of measurements of system variables is taken until the actual faults are identified. The goal is to reduce the diagnostic cost, defined here as the number of measurements. To propose measurement points, the state-of-the-art GDE (general diagnosis engine) framework (de Kleer & Williams, 1987; de Kleer, Raiman, & Shirley, 1992; de Kleer, 2006) considers a heuristic based on reducing the entropy over a set of computed diagnoses. This approach generally has good performance in terms of diagnostic cost, but can fail to diagnose large systems when the set of diagnoses is too large (de Kleer & Williams, 1987; de Kleer et al., 1992; de Kleer, 2006). Focusing on a smaller set of *probable diagnoses* scales the approach but generally leads to increased average diagnostic costs (de Kleer, 1992).

We propose a new diagnostic framework employing four new techniques, which scales to much larger systems with good performance in terms of diagnostic cost. First, we propose a new heuristic that does not require computing the entropy of diagnoses. Instead we consider the entropies of the system variables to be measured as well as the posterior probabilities of component failures. The idea is to select a component that has the highest posterior probability of failure (Heckerman, Breese, & Rommelse, 1995) and from the variables of that component, measure the one that has the highest entropy. To compute probabilities, we exploit system structure so that a joint probability distribution over the faults and system variables is represented compactly as a *Bayesian network* (Pearl, 1988), which is then compiled into *deterministic decomposable negation normal form* (d-DNNF) (Darwiche, 2001; Darwiche & Marquis, 2002). d-DNNF is a logical form that can exploit the structure present in many systems to achieve compactness and be used to compute probabilistic queries efficiently. Specifically, all the required posterior probabilities can be exactly computed by evaluating and differentiating the d-DNNF in time linear in the d-DNNF size (Darwiche, 2003).

Second, we extend *hierarchical diagnosis*, a technique from our previous work (Siddiqi & Huang, 2007), to handle probabilities so that it can be applied to sequential diagnosis to allow larger systems to be diagnosed. Specifically, self-contained subsystems, called *cones*, are treated as single components and diagnosed only if they are found to be faulty in the top-level diagnosis. This significantly reduces the number of system components, allowing larger systems to be compiled and diagnosed. For example, the subcircuit in the dotted box in Figure 1 is a cone (with $A$ as output and $\{P, D\}$ as inputs) which contains a fault. First, cone $A$, as a whole, is determined as faulty. It is only then that $A$ is compiled separately and diagnosed. In previous work (Siddiqi & Huang, 2007) we only dealt with the task of computing diagnoses, which did not involve measurements or probabilities; in the present paper, we present several extensions that allow the technique to carry over to sequential diagnosis.

Third, when the abstraction of a system is still too large to be compiled and diagnosed, we use a novel structure based technique called *cloning*, which systematically modifies the structure of a given system $\mathbf{C}$ to obtain a new system $\mathbf{C}'$ that has a smaller abstraction and whose diagnoses form a super-set of those of the original system; the new system can then be diagnosed and the result mapped back to the original system. The idea is to select a system component $G$ that is not part of a cone and hence cannot be abstracted away in hierarchical diagnosis, create one or more clones of $G$, and distribute $G$'s parents (from a graph point of view) among the clones, in such a way that $G$ and its clones now become parts of cones and disappear from the abstraction. Repeated applications of this operation can allow an otherwise unmanageable system to have a small enough abstraction for diagnosis to succeed.

Finally, we propose a novel cost estimation function that can predict the expected diagnostic cost when a given abstraction of the system is used for diagnosis. Our aim is to find an abstraction of the system that is more likely to give optimal average cost. For this purpose, we use this function on various abstractions of the system where different abstractions are obtained by destroying different cones in the system (by "destroying a cone" we mean to overlook the fact that it is a cone and include all its components in the abstraction). The abstraction with the lowest predicted cost can then be used for the actual diagnosis.

Experiments on ISCAS-85 benchmark circuits (Brglez & Fujiwara, 1985) indicate that we can solve for the first time nontrivial multiple-fault diagnostic cases on all the benchmarks, with good diagnostic costs, except one circuit that has a flat structure not susceptible to useful abstraction, and the new cost estimation function can often accurately predict the abstraction which is more likely to give optimal average cost.

## 2. Background and Previous Work

Suppose that the system to be diagnosed is formally modeled by a joint probability distribution $Pr(\mathbf{X} \cup \mathbf{H})$ over a set of variables partitioned into $\mathbf{X}$ and $\mathbf{H}$. Variables $\mathbf{X}$ are those whose values can be either observed or measured, and variables $\mathbf{H}$ are the *health* variables, one for each component describing its health mode. The joint probability distribution $Pr(\mathbf{X} \cup \mathbf{H})$ defines a set of system states.

Diagnosis starts in the initial (belief) state

$$I_0 = Pr(\mathbf{X} \cup \mathbf{H} \mid \mathbf{X_o} = \mathbf{x_o}) \tag{1}$$

where values $\mathbf{x_o}$ of some variables $\mathbf{X_o} \subseteq \mathbf{X}$ (we are using boldface uppercase letters to mean both sets and vectors) are given by the observation, and we wish to reach a goal state

$$I_n = Pr(\mathbf{X} \cup \mathbf{H} \mid \mathbf{X_o} = \mathbf{x_o}, \mathbf{X_m} = \mathbf{x_m}) \tag{2}$$

after measuring the values $\mathbf{x_m}$ of some variables $\mathbf{X_m} \subseteq \mathbf{X} \backslash \mathbf{X_o}$, $|\mathbf{X_m}| = n$, one at a time, such that (the boldface $\mathbf{0}$ and $\mathbf{1}$ denote vectors of 0's and 1's):

$$\exists \mathbf{H_f} \subseteq \mathbf{H}, Pr(\mathbf{H_f} = \mathbf{0} \mid \mathbf{X_o} = \mathbf{x_o}, \mathbf{X_m} = \mathbf{x_m}) = 1 \text{ and}$$

$$Pr(\mathbf{H_f} = \mathbf{0}, \mathbf{H} \backslash \mathbf{H_f} = \mathbf{1} \mid \mathbf{X_o} = \mathbf{x_o}, \mathbf{X_m} = \mathbf{x_m}) > 0.$$

That is, in a goal state a set of components $\mathbf{H_f}$ are known to be faulty with certainty and no logical inconsistency arises if all other components are assumed to be healthy. Other types of goal conditions are possible. For example, if the health states of all components are to be determined with certainty, the condition will be that $Pr(H = 0 \mid \mathbf{X_o} = \mathbf{x_o}, \mathbf{X_m} = \mathbf{x_m})$ is 0 or 1 for all $H \in \mathbf{H}$ (such goals are only possible to reach if *strong fault models* are given, where strong fault models are explicit descriptions of abnormal behavior, as opposed to *weak fault models* where only the normal behavior is known).

Two special cases are worth mentioning: (1) If the initial state $I_0$ satisfies the goal condition with $\mathbf{H_f} = \emptyset$ then the observation is normal and no diagnosis is required. (2) If the initial state $I_0$ satisfies the goal condition with some $\mathbf{H_f} \neq \emptyset$, then the observation is abnormal but the diagnosis is already completed (assuming that we are able to check probabilities as necessary); in other words, a sequence of length 0 solves the problem.

Following de Kleer and Williams (1987) we assume that all measurements have unit cost. Hence the objective is to reach a goal state in the fewest measurements possible.

The classical GDE framework, on receiving an abnormal observation $\mathbf{X_o} = \mathbf{x_o}$, considers the Shannon's entropy of the probability distribution over a set of computed diagnoses, which is either the set of *minimum-cardinality diagnoses* or a set of *probable/leading diagnoses*. It proposes to measure a variable $X$ whose value will reduce that entropy the most, on average. The idea is that the probability distribution over the diagnoses reflects the uncertainty over the actual faults, and the entropy captures the amount of this uncertainty. After a measurement is taken the entropy is updated by updating the posterior probabilities of the diagnoses, potentially reducing some of them to 0.

The results reported by de Kleer et al. (1992) involving single-fault cases for ISCAS-85 circuits indicate that this method leads to measurement costs close to those of optimal policies. However, a major drawback is that it can be impractical when the number of diagnoses is large (e.g., the set of minimum-cardinality diagnoses can be exponentially large). Focusing on a smaller set of probable diagnoses scales the approach but can increase the likelihood of irrelevant measurements and generally leads to increased average diagnostic costs (de Kleer, 1992).

From here on, we shall use combinational circuits as an example of the type of systems we wish to diagnose. Our approach, however, applies as well to other types of systems as long as a probabilistic model is given that defines the behavior of the system. In Sections 4 and 5 we will present the new techniques we have introduced to significantly enhance the scalability of sequential diagnosis. We start, however, by presenting in the following section the system modeling and compilation method that underlies our new diagnostic system.

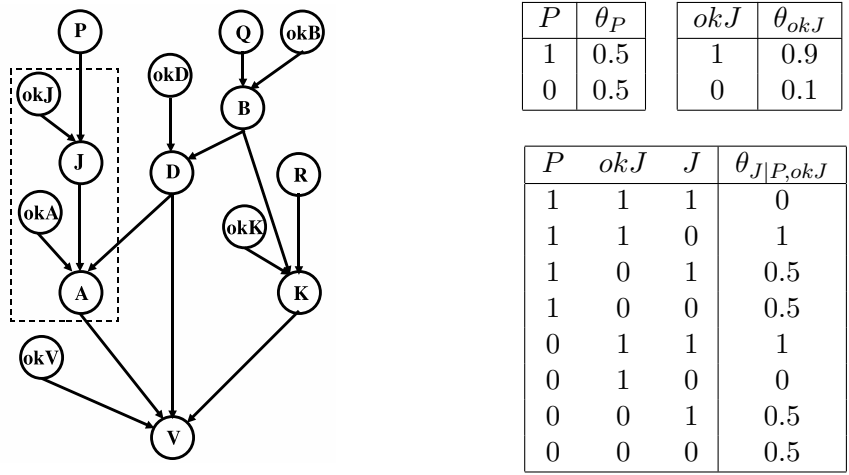

Figure 2: Bayesian network for the circuit in Figure 1 (left). CPTs for nodes $P$, $J$, and $okJ$ (right).

## 3. System Modeling and Compilation

In order to define a joint probability distribution $Pr(\mathbf{X} \cup \mathbf{H})$ over the system behavior, we first assume that the prior probability of failure $Pr(H = 0)$ is given for each component $H \in \mathbf{H}$ as part of the input to the diagnosis task (de Kleer & Williams, 1987). For example, the small table with two entries on the top-right of Figure 2 gives the prior probability of failure for gate $J$ as 0.1.

### 3.1 Conditional Probability Tables

Prior fault probabilities alone do not define the joint probability distribution $Pr(\mathbf{X} \cup \mathbf{H})$. In addition, we need to specify for each component how its output is related to its inputs and health mode. A conditional probability table (CPT) for each component does this job.

The CPT shown on the bottom (right) of Figure 2, for example, defines the behavior of gate $J$: Each entry gives the probability of its output ($J$) being a particular value given the value of its input ($P$) and the value of its health variable ($okJ$). In case $okJ = 1$, the probabilities are always 0 or 1 as the behavior of a healthy gate is deterministic. The case of $okJ = 0$ defines the *fault model* of the gate, which is also part of the input to the diagnosis task. In our example, we assume that both output values have probability 0.5 when the gate is broken. For simplicity we assume that all gates have two health modes (i.e., each health variable is binary); the encoding and compilation to be described later, however, allows an arbitrary number of health modes.

Given these tables, the joint probability distribution over the circuit behavior can be obtained by realizing that the gates of a circuit satisfy an independence property, known as the *Markov property*: Given its inputs and health mode, the output of a gate is independent of any wire which is not a descendant of the gate (a wire $X$ is a descendant of a gate $Y$ if $X$ can be reached following a path from $Y$ to an output of the circuit in the direction towards the circuit outputs). This means that the circuit can be effectively treated as a Bayesian network in the straightforward way, by having a node for each wire and each health variable, and having an edge going from each input of a gate to its output, and also from the health variable of a gate to its output. Figure 2 shows the result of this translation for the circuit in Figure 1.

The joint probability distribution encoded in the Bayesian network provides the basis for computing any posterior probabilities that we may need when proposing measurement points (by the *chain rule*). However, it does not provide an efficient way of doing so. Specifically, computing a posterior $Pr(X = x \mid \mathbf{Y} = \mathbf{y})$ given the values $\mathbf{y}$ of all the variables $\mathbf{Y}$ with known values involves summing out all variables other than $X$ and $\mathbf{Y}$, which has a complexity exponential in the number of such variables if done naively.

## 3.2 Propositional Modeling

It is known that a Bayesian network can be encoded into a logical formula and compiled into d-DNNF, which, if successful, allows posterior probabilities of all variables to be computed efficiently (Darwiche, 2003). For the purposes of sequential diagnosis, we encode the Bayesian network as follows.

Consider the subcircuit in the dotted box in Figure 1 as an example, which can be modeled as the following formula:

$$okJ \rightarrow (J \leftrightarrow \neg P), \;\; okA \rightarrow (A \leftrightarrow (J \wedge D)).$$

Specifically, each signal of the circuit translates into a propositional variable ($A$, $D$, $P$, $J$), and for each gate, an extra variable is introduced to model its health ($okA$, $okJ$). The formula is such that when all health variables are *true* the remaining variables are constrained to model the functionality of the gates. In general, for each component $X$, we have $okX \rightarrow \textsc{NormalBehavior}(X)$.

Note that the above formula fails to encode half of the CPT entries, where $okJ = 0$. In order to complete the encoding of the CPT of node $J$, we introduce an extra Boolean variable $\theta_J$, and write $\neg okJ \rightarrow (J \leftrightarrow \theta_J)$. Finally, the health variables ($okA$, $okJ$) are associated with the probabilities of the respective gates being healthy (0.9 in our experiments), and each $\theta$-variable ($\theta_J$) is associated with the probability of the corresponding gate giving an output of 1 when broken (0.5 in our experiments; thus assuming that the output of a faulty gate is probabilistically independent of its inputs).

The above encoding of the circuit is similar to the encoding of Bayesian networks described by Darwiche (2003) in the following way: According to the encoding by Darwiche, for every node in a Bayesian network and for every value of it there is an *indicator* variable. Similarly for every conditional probability there is a *network parameter* variable. In our encoding, the variables for the wires are analogous to the network indicators, where the encoding is optimized such that there is a single indicator for both values of the wire. Also, our encoding exploits the *logical constraints* and does not generate network parameters for zeros and ones in the CPT. Finally, the encoding for a node that represents a health variable has been optimized such that we only need a single *ok*-variable which serves both as an indicator and as a network parameter.

Once all components are encoded as described above, the union (conjunction) of the formulas is compiled into d-DNNF. The required probabilities can be exactly computed by evaluating and differentiating the d-DNNF in time linear in its size (Darwiche, 2003). Details of the compilation process are discussed by Darwiche (2004), and the computation of probabilities is described in Appendix A.

We now present our hierarchical diagnosis approach and propose a new measurement selection heuristic.

## 4. Hierarchical Sequential Diagnosis

An optimal solution to sequential diagnosis would be a *policy*, that is, a plan of measurements conditioned on previous measurement outcomes, where each path in the plan leads to a diagnosis of the system (Heckerman et al., 1995). As computing optimal policies is intractable in general, we follow the approach of heuristic measurement point selection as in previous work.

We start with a definition of Shannon's entropy $\xi$, which is defined with respect to a probability distribution of a discrete random variable $X$ ranging over values $x_1, x_2, \ldots, x_k$. Formally:

$$\xi(X) = -\sum_{i=1}^{k} Pr(X = x_i) \log Pr(X = x_i). \tag{3}$$

Entropy measures the amount of uncertainty over the value of the random variable. It is maximal when all probabilities $Pr(X = x_i)$ are equal, and minimal when one of the probabilities is 1, corresponding nicely to our intuitive notion of the degree of uncertainty. In GDE the entropy is computed for the probability distribution over the set of computed diagnoses (i.e., the value of the random variable $X$ here ranges over the set of diagnoses). As mentioned earlier, this entropy can be difficult to compute when the number of diagnoses is large (de Kleer & Williams, 1987; de Kleer, 2006).

### 4.1 Baseline Approach

Able to compute probabilities efficiently and exactly following successful d-DNNF compilation, we now propose a new two-part heuristic that circumvents this limitation in scalability. First, we consider the entropy of a candidate variable to be measured.

#### 4.1.1 Heuristic Based on Entropy of Variable

Since a wire $X$ only has two values, its entropy can be written as:

$$\xi(X) = -(p_x \log p_x + p_{\bar{x}} \log p_{\bar{x}}) \tag{4}$$

where $p_x = Pr(X = 1 \mid \mathbf{Y} = \mathbf{y})$ and $p_{\bar{x}} = Pr(X = 0 \mid \mathbf{Y} = \mathbf{y})$ are the posterior probabilities of $X$ having values 1 and 0, respectively, given the values $\mathbf{y}$ of wires $\mathbf{Y}$ whose values are known.

While $\xi(X)$ captures the uncertainty over the value of the variable, we can also interpret it as the expected amount of information gain provided by measuring the variable. Hence as a first idea we consider selecting a variable with maximal entropy for measurement at each step.

### 4.1.2 Improving Heuristic Accuracy

This idea alone, however, did not work very well in our initial experiments. As would be confirmed by subsequent experiments, this is largely due to the fact that the (implicit) space of all diagnoses is generally very large and can include a large number of unlikely diagnoses, which tends to compromise the accuracy of the information gain provided by the entropy. The experiments to confirm this explanation are as follows.

When the d-DNNF compilation is produced, and before it is used to compute probabilities, we prune the d-DNNF graph so that models (satisfying variable assignments) corresponding to diagnoses with more than $k$ broken components are removed.[1] We set the initial $k$ to the number of actual faults in the experiments, and observed that a significant reduction of diagnostic cost resulted in almost all cases. This improved performance is apparently due to the fact that the pruning updates the posterior probabilities of all variables, making them more accurate since many unlikely diagnoses have been eliminated.

In practice, however, the number of faults is not known beforehand and choosing an appropriate $k$ for the pruning can be nontrivial (note that $k$ need not be exactly the same as the number of actual faults for the pruning to help). Interestingly, the following heuristic, which is the one we will actually use, appears to achieve a similar performance gain in an automatic way: We select a component that has the highest posterior probability of failure (an idea from Heckerman et al., 1995; see Section 8), and then from the variables of that component, measure the one that has the highest entropy. This heuristic does not require the above pruning of the d-DNNF, and appears to improve the diagnostic cost to a similar extent by focusing the measurement selection on the component most likely to be broken (empirical results to this effect are given and discussed in Section 7.1).

### 4.1.3 The Algorithm

We start by encoding the system as a logical formula as discussed in Section 3, where a subset of the variables are associated with numbers representing the prior fault probabilities and probabilities involved in the fault models of the components, which is then compiled into d-DNNF $\Delta$.

The overall sequential diagnosis process we propose is summarized in Algorithm 1. The inputs are a system $\mathbf{C}$, its d-DNNF compilation $\Delta$, the set of faults $\mathbf{D}$ (which is empty but will be used in the hierarchical approach), a set of known values $\mathbf{y}$ of variables, and an integer $k$ specifying the fault cardinality bound (this is for running the model pruning experiments described in Section 4.1.2, and is not required for diagnosis using our final

---

1. A complete pruning is not easy; however, an approximation can be achieved in time linear in the d-DNNF size, by a variant of the minimization procedure described by Darwiche (2001); see Appendix B.

---

**Algorithm 1** Probabilistic sequential diagnosis

---

**function** PSD($\mathbf{C}$, $\Delta$, $\mathbf{D}$, $\mathbf{y}$, $k$)

**inputs:** {$\mathbf{C}$: system}, {$\Delta$: d-DNNF}, {$\mathbf{y}$: measurements}, {$k$: fault cardinality}, {$\mathbf{D}$: ordered set of known faults}

**output:** {pair< $\mathbf{D}$ , $\mathbf{y}$ >}

1: $\Delta \leftarrow$ REDUCE ( $\Delta$, $\mathbf{D}$, $k - |\mathbf{D}|$ ) if $\mathbf{D}$ has changed
2: Given $\mathbf{y}$ on variables $\mathbf{Y}$, EVALUATE ($\Delta$, $\mathbf{y}$) to obtain $Pr(\mathbf{y})$
3: DIFFERENTIATE ($\Delta$) to obtain $Pr(X = 1, \mathbf{y})$ $\forall$ variables $X$
4: Deduce fault as $\mathbf{D} = \mathbf{D} \cup \{X : Pr(okX = 1, \mathbf{y}) = 0\}$
5: **if** $\mathbf{D}$ has changed && MEETSCRITERIA($\Delta$,$\mathbf{D}$,$\mathbf{y}$) **then**
6:     **return** < $\mathbf{D}$ , $\mathbf{y}$ >
7: Measure variable $X$ which is the best under a given heuristic
8: Add the measured value $x$ of $X$ to $\mathbf{y}$, and go back to line 1

---

heuristic). We reduce $\Delta$ by pruning some models (line 1) when the fault cardinality bound $k$ is given, using the function REDUCE($\Delta$, $\mathbf{D}$, $k - |\mathbf{D}|$). REDUCE accepts as arguments the current DNNF $\Delta$, the set of known faults $\mathbf{D}$, and the upper bound given by $k - \mathbf{D}$ on the cardinality of remaining faults, whereas it returns the pruned DNNF. REDUCE excludes the known faults in $\mathbf{D}$ when computing the minimum cardinality of $\Delta$, and then uses $k - |\mathbf{D}|$ as the bound on the remaining faults (explained further in Appendix B). $\Delta$ is reduced first time when PSD is called and later each time $\mathbf{D}$ is changed (i.e., when a component is found faulty). We then evaluate (line 2) and differentiate (line 3) $\Delta$ (see Appendix A), select a measurement point and take the measurement (line 7), and repeat the process (line 8) until the stopping criteria are met (line 5).

The stopping criteria on line 5 are given earlier in Section 2 as the goal condition, i.e., we stop when the abnormal observation is explained by all the faulty components $\mathbf{D}$ already identified assuming that other components are healthy. A faulty component $X$ is identified when $Pr(okX = 1, \mathbf{y}) = 0$ where $\mathbf{y}$ are the values of variables that are already known, and as mentioned earlier these probabilities are obtained for all variables simultaneously in the d-DNNF differentiation process. Finally, the condition that the current set of faulty components, with health modes $\mathbf{H_f}$, explains the observation is satisfied when $Pr(\mathbf{H_f} = \mathbf{0}, \mathbf{H} \backslash \mathbf{H_f} = \mathbf{1}, \mathbf{y}) > 0$, which is checked by a single evaluation of the original d-DNNF. The algorithm returns the actual faults together with the new set of known values of variables (line 6).

## 4.2 Hierarchical Approach

We now scale our approach to handle larger systems using the idea of abstraction-based hierarchical diagnosis (Siddiqi & Huang, 2007). The basic idea is that the compilation of the system model into d-DNNF will be more efficient and scalable when the number of system components is reduced. This can be achieved by abstraction, where subsystems, known as cones, are treated as single components. An example of a cone is depicted in Figure 1. The objective here is to use a single health variable and failure probability for the entire cone, hence significantly reducing the size of the encoding and the difficulty of compilation. Once a cone is identified as faulty in the top-level diagnosis, it can then be compiled and diagnosed, in a recursive fashion.

We now give formal definition of abstraction from our previous work:

### 4.2.1 Abstraction of System

Abstraction is based upon the *structural dominators* (Kirkland & Mercer, 1987) of a system. A component $X$ dominates a component $Y$, or $X$ is called a dominator of $Y$, if any path from $Y$ to any output of the system contains $X$. A cone corresponds precisely to the set of components dominated by a component. A cone may contain further cones leading to a hierarchy of cones.

A system can be abstracted by treating all maximal cones in it as black boxes (a maximal cone is one that is either contained in no other cone or contained in exactly one other cone which is the whole system). In our example, cone $A$ can be treated as a virtual gate with two inputs $\{P, D\}$ and the output $A$. The abstraction of a system can be formally defined as:

**Definition 1** (Abstraction of System). *Given a system* $\mathbf{C}$, *let* $\mathbf{C}' = \mathbf{C}$ *if* $\mathbf{C}$ *has a single output; otherwise let* $\mathbf{C}'$ *be* $\mathbf{C}$ *augmented with a dummy component collecting all outputs of* $\mathbf{C}$. *Let* $O$ *be the only output of* $\mathbf{C}'$. *The abstraction* $\mathbf{A_C}$ *of system* $\mathbf{C}$ *is then the set of components* $X \in \mathbf{C}$ *such that* $X$ *is not dominated in* $\mathbf{C}'$ *by any component other than* $X$ *and* $O$.

For example, $\mathbf{A_C} = \{A, B, D, K, V\}$. $J \notin \mathbf{A_C}$ as $J$ cannot reach any output without passing through $A$, which is a dominator of $J$.

In our previous work (Siddiqi & Huang, 2007), we only dealt with the task of computing minimum-cardinality diagnoses, which does not involve probabilities or measurement selection. In the context of sequential diagnosis, several additional techniques have been introduced, particularly in the computation of prior failure probabilities for the cones and the way measurement points are selected, outlined below.

### 4.2.2 Propositional Encoding

We start with a discussion of the hierarchical encoding for probabilistic reasoning, which is similar to the hierarchical encoding presented in our previous work (Siddiqi & Huang, 2007). Specifically, for the diagnosis of the abstraction $\mathbf{A_C}$ of the given system $\mathbf{C}$, health variables are only associated with the components $\mathbf{A_C} \backslash \mathbf{I_C}$, which are the gates $\{A, B, D, K, V\}$ in our example ($\mathbf{I_C}$ stands for the set of inputs of the system $\mathbf{C}$). Thus the gate $J$ in Figure 1 will not be associated with a health variable, as $J$ is a wire internal to the cone rooted at $A$. Consequently, only the nodes representing the components $\mathbf{A_C} \backslash \mathbf{I_C}$ will have health nodes associated with them in the corresponding Bayesian network. Hence the node $okJ$ is removed from the Bayesian network in Figure 2.

In addition, we define the failure of a cone to be when it outputs the wrong value, and introduce extra clauses to model the abnormal behavior of the cone. For example, the encoding given in Section 3.2 for cone $A$ in Figure 1 (in the dotted box) is as follows:

$$J \leftrightarrow \neg P, \;\; okA \rightarrow (A \leftrightarrow (J \wedge D)), \;\; \neg okA \rightarrow (A \not\leftrightarrow (J \wedge D))$$

The first part of the formula encodes the normal behavior of gate $J$ (without a health variable); the next encodes the normal behavior of the cone; the last encodes that the cone outputs a wrong value when it fails. Other gates (that are not roots of cones) in the abstraction $\mathbf{A_C}$ are encoded normally as described in Section 3.2.

Note that the formulas for all the components in a cone together encode a single CPT for the whole cone, which provides the conditional probability of the cone's output given the health and inputs of the cone, instead of the health and inputs of the component at the root of the cone. For example, the above encoding is meant to provide the conditional probability of $A$ given $P$, $D$, and $okA$ (instead of $J$, $D$, and $okA$), where $okA$ represents the health mode of the whole cone and is associated with its prior failure probability, which is initially unknown to us and has to be computed for all cones (explained below). Such an encoding of the whole system provides a joint probability distribution over the variables $\mathbf{A_C} \cup \mathbf{I_C} \cup \mathbf{H}$, where $\mathbf{H} = \{okX \mid X \in \mathbf{A_C} \backslash \mathbf{I_C}\}$.

### 4.2.3 Prior Failure Probabilities for Cones

When a cone is treated as a single component, its prior probability of failure as a whole can be computed given the prior probabilities of components and cones inside it. We do this by creating two copies $\Delta_h$ and $\Delta_f$ of the cone, where $\Delta_h$ models only the healthy behavior of the cone (without health variables), and $\Delta_f$ includes the faulty behavior as well (i.e., the full encoding described in Section 3.2). The outputs of both $\Delta_h$ and $\Delta_f$ are collected into an XOR-gate $X$(when the output of XOR-gate $X$ equals 1, both of its inputs are forced to be different in value). We then compute the probability $Pr(X = 1)$ giving the probability of the outputs of $\Delta_h$ and $\Delta_f$ being different. The probability is computed by compiling this encoding into d-DNNF and evaluating it under $X = 1$.

Note that this procedure itself is also abstraction-based and hierarchical, performed bottom-up with the probabilities for the inner cones computed before those for the outer ones. Also note that it is performed only once per system as a pre-processing step.

### 4.2.4 Measurement Point Selection and Stopping Criteria

In principle, the heuristic to select variables for measurement and the stopping criteria are the same as in the baseline approach; however, a couple of details are worth mentioning.

First, when diagnosing the abstraction of a given system (or cone) $\mathbf{C}$, the measurement candidates are restricted to variables $\mathbf{A_C} \cup \mathbf{I_C}$, ignoring the internal variables of the maximal cones—those are only measured if a cone as a whole has been found faulty.

Second, it is generally important to have full knowledge of the values of cone's inputs before a final diagnosis of the cone is concluded. A diagnosis of a cone concluded with only partial knowledge of its inputs may not include some faults that are vital to the validity of global diagnosis. The reason is that the diagnosis of the cone assumes that the unknown inputs can take either value, while in reality their values may become fixed when variables in other parts of the system are measured, causing the diagnosis of certain cones to become invalid, and possibly requiring the affected cones to be diagnosed once again to meet the global stopping criteria (see line 17 in Algorithm 2).

To avoid this situation while retaining the effectiveness of the heuristic, we modify the measurement point selection as follows when diagnosing a cone. After selecting a component with the highest probability of failure, we consider the variables of that component *plus* the inputs of the cone, and measure the one with the highest entropy. We do not conclude a

---

**Algorithm 2** Hierarchical probabilistic sequential diagnosis

---

**function** HPSD($\mathbf{C}$, $\mathbf{u_C}$, $k$)
**inputs:** $\{\mathbf{C}$ : system$\}$,$\{\mathbf{u_C}$: obs. across system$\}$ $\{k$: fault cardinality$\}$
**local variables:** $\{\mathbf{B}, \mathbf{D}, \mathbf{T}$ : set of components$\}$ $\{\mathbf{y}, \mathbf{z}, \mathbf{u_G}$ : set of measurements$\}$ $\{i, k'$ : integer$\}$
**output:** $\{$pair$< \mathbf{D}$ , $\mathbf{u_C} >\}$
1: $\Delta \leftarrow$ COMPILE2DDNNF ($\mathbf{A_C}$, $\mathbf{u_C}$)
2: $i \leftarrow 0$ , $\mathbf{D} \leftarrow \phi$ , $\mathbf{y} \leftarrow \mathbf{u_C}$
3: $< \mathbf{B}, \mathbf{y} > \leftarrow$ PSD ($\mathbf{C}$, $\Delta$, $\mathbf{B}$, $\mathbf{y}$, $k$)
4: **for** $\{; i < |\mathbf{B}|; i++\}$ **do**
5: $\quad$ $G \leftarrow$ ELEMENT ($\mathbf{B}$, $i$)
6: $\quad$ **if** $G$ is a cone **then**
7: $\quad\quad$ $\mathbf{z} \leftarrow \mathbf{y} \cup$ IMPLICATIONS ($\Delta$, $\mathbf{y}$)
8: $\quad\quad$ $\mathbf{u_G} \leftarrow \{x : x \in \mathbf{z}, X \in \mathbf{I_G} \cup \mathbf{O_G}\}$
9: $\quad\quad$ $k' \leftarrow k - |\mathbf{D}| - |\mathbf{B}| + i + 2$
10: $\quad\quad$ $< \mathbf{T}, \mathbf{u_G} > \leftarrow$ HPSD($\mathbf{D_G} \cup \mathbf{I_G}$, $\mathbf{u_G}$, $k'$)
11: $\quad\quad$ $\mathbf{y} \leftarrow \mathbf{y} \cup \mathbf{u_G}$ , $\mathbf{D} \leftarrow \mathbf{D} \cup \mathbf{T}$
12: $\quad\quad$ EVALUATE ($\Delta$, $\mathbf{y}$), DIFFERENTIATE ( $\Delta$ )
13: $\quad$ **else**
14: $\quad\quad$ $\mathbf{D} \leftarrow \mathbf{D} \cup \{G\}$
15: $\mathbf{z} \leftarrow \mathbf{y} \cup$ IMPLICATIONS ($\Delta$, $\mathbf{y}$)
16: $\mathbf{u_C} \leftarrow \mathbf{u_C} \cup \{x : x \in \mathbf{z}, X \in \mathbf{I_C} \cup \mathbf{O_C}\}$
17: **if** MEETSCRITERIA ($\mathbf{C}$, $\mathbf{D}$, $\mathbf{y}$) **then**
18: $\quad$ **return** $< \mathbf{D}$ , $\mathbf{u_C} >$
19: **else**
20: $\quad$ goto line 3

---

diagnosis for the cone until values of all its inputs become known (through measurement or deduction), except when the health of all the components in the cone has been determined without knowing all the inputs to the cone (it is possible to identify a faulty component, and with strong fault models also a healthy component, without knowing all its inputs). Note that the restriction of having to measure all the inputs of a cone can lead to significant increase in the cost compared with the cost of baseline approach; especially when the number of inputs of a cone is large. This is discussed in detail in Section 6.

### 4.2.5 THE ALGORITHM

Pseudocode for the hierarchical approach is given in Algorithm 2 as a recursive function. The inputs are a system $\mathbf{C}$, a set of known values $\mathbf{u_C}$ of variables at the inputs $\mathbf{I_C}$ and outputs $\mathbf{O_C}$ of the system, and again the optional integer $k$ specifying the fault cardinality bound for the purpose of experimenting with the effect of model pruning. We start with the d-DNNF compilation of the abstraction of the given system (line 1) and then use the function PSD from Algorithm 1 to get a diagnosis $\mathbf{B}$ of the abstraction (line 3), assuming that the measurement point selection and stopping criteria in Algorithm 1 have been modified according to what is described in Section 4.2.4. The abstract diagnosis $\mathbf{B}$ is then used to get a concrete diagnosis $\mathbf{D}$ in a loop (lines 4–14). Specifically, if a component $G \in \mathbf{B}$ is not the root of a cone, then it is added to $\mathbf{D}$ (line 14); otherwise cone $G$ is recursively diagnosed (line 10) and the result of it added to $\mathbf{D}$ (line 11). When recursively diagnosing a cone $G$, the subsystem contained in $G$ is represented by $\mathbf{D_G} \cup \mathbf{I_G}$, where $\mathbf{D_G}$ is the set of components dominated by $G$ and $\mathbf{I_G}$ is the set of inputs of cone $G$.

Before recursively diagnosing a cone $G$, we compute an abnormal observation $\mathbf{u_G}$ at the inputs and the output ($\mathbf{I_G} \cup \{G\}$) of the cone $G$. The values of some of $G$'s inputs and output will have been either measured or deduced from the current set of measurements. The value of a variable $X$ is implied to be $x$ under the measurements $\mathbf{y}$ if $Pr(X = \neg x, \mathbf{y}) = 0$, which is easy to check once $\Delta$ has been differentiated under $\mathbf{y}$. The function IMPLICATIONS($\Delta, \mathbf{y}$) (lines 7 and 15) implements this operation, which is used to compute the partial abnormal observation $\mathbf{u_G}$ (line 8). A fault cardinality bound $k'$ for the cone $G$ is then inferred (line 9), and the algorithm called recursively to diagnose $G$, given $\mathbf{u_G}$ and $k'$.

The recursive call returns the faults $\mathbf{T}$ inside the cone $G$ together with the updated observation $\mathbf{u_G}$. The observation $\mathbf{u_G}$ may contain some new measurement results regarding the variables $\mathbf{I_G} \cup \{G\}$, which are added to the set of measurements $\mathbf{y}$ of the abstraction (line 11); other measurement results obtained inside the cone are ignored due to reasons explained in Section 4.2.4. The concrete diagnosis $\mathbf{D}$ is augmented with the faults $\mathbf{T}$ found inside the cone (line 11), and $\Delta$ is again evaluated and differentiated in light of the new measurements (line 12).

After the loop ends, the variable $\mathbf{u_C}$ is updated with the known values of the inputs $\mathbf{I_C}$ and outputs $\mathbf{O_C}$ of the system $\mathbf{C}$ (line 16). The stopping criteria are checked for the diagnosis $\mathbf{D}$ (line 17) and if met the function returns the pair $< \mathbf{D}, \mathbf{u_C} >$ (line 18); otherwise more measurements are taken until the stopping criteria (line 17) have been met.

Since $\mathbf{D}$ can contain faults from inside the cones, the compilation $\Delta$ cannot be used to check the stopping criteria for $\mathbf{D}$ (note the change in the parameters to the function MEETSCRITERIA at line 17) as the probabilistic information regarding variables inside cones is not available in $\Delta$. The criteria are checked as follows instead: We maintain the depth level of every component in the system. The outputs of the system are at depth level 1 and the rest of the components are assigned depth levels based upon the length of their shortest route to an output of the system. For example, in Figure 1 gates $B$ and $J$ are at depth level 3, while $A$ is at depth level 2. Hence, $B$ and $J$ are deeper than $A$. We first propagate the values of inputs in the system, and then propagate the fault effects of components in $\mathbf{D}$, one by one, by flipping their values to the abnormal ones and propagating them towards the system outputs in such a way that deeper faults are propagated first (Siddiqi & Huang, 2007), and then check the values of system outputs obtained for equality with those in the observation ($\mathbf{y}$).

### 4.2.6 Example

Suppose that we diagnose the abstraction of the circuit in Figure 1, with the observation $\mathbf{u_C} = \{P = 1, Q = 1, R = 0, V = 1\}$, and take the sequence of measurements $\mathbf{y} = \{D = 1, K = 1, A = 1\}$. It is concluded, from the abstract system model, that given the values of $P$ and $D$, the value 1 at $A$ is abnormal. So the algorithm concludes a fault at $A$. Note that $Q = 1$ and $D = 1$ suggests the presence of another fault besides $A$, triggering the measurement of gate $B$, which is also found faulty. The abstract diagnosis $\{A, B\}$ meets the stopping criteria with respect to the abstract circuit.

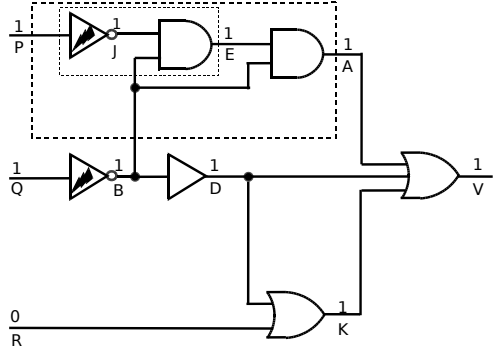

Figure 3: A faulty circuit with faults at $B$ and $J$.

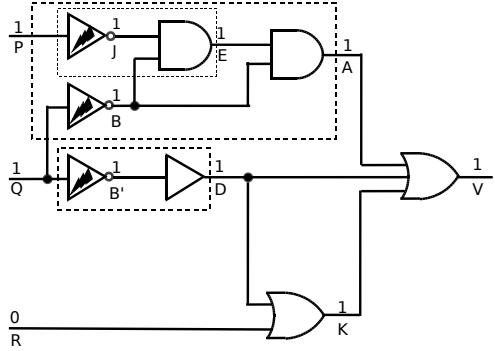

Figure 4: Creating a clone $B'$ of $B$ according to $D$.

We then enter the diagnosis of cone $A$ by a recursive call with observation $\mathbf{u_A} = \{P = 1, B = 1, A = 1\}$. The diagnosis of the cone $A$ immediately reveals that the cone $E$ is faulty. Hence we make a further recursive call in order to diagnose $E$ with the observation $\mathbf{u_E} = \{P = 1, B = 1, E = 1\}$. The only unknown wire $J$ is measured and the gate $J$ is found faulty, which explains the observation at the outputs of the cones $E$ as well as $A$, given the inputs $P$ and $B$. The recursion terminates and the abstract diagnosis $\mathbf{B} = \{A, B\}$ generates the concrete diagnosis $\mathbf{D} = \{J, B\}$, which meets the stopping criteria and the algorithm terminates.

## 5. Component Cloning

In the preceding section, we have proposed an abstraction-based approach to sequential diagnosis, which reduces the complexity of compilation and diagnosis by reducing the number of system components to be diagnosed. We now take one step further, aiming to handle systems that are so large that they remain intractable even after abstraction, as is the case for the largest circuits in the ISCAS-85 benchmark suite.

Our solution is a novel method that systematically modifies the structure of a system to reduce the size of its abstraction. Specifically, we select a component $G$ with parents $\mathbf{P}$ (a component $X$ is a *parent* of a component $Y$, and $Y$ is a *child* of $X$, if the output of $Y$ is an input of $X$) that is not part of a cone and hence cannot be abstracted away in hierarchical diagnosis, and create a clone $G'$ of it according to some of its parents $\mathbf{P}' \subset \mathbf{P}$ in the sense that $G'$ inherits all the children of $G$ and feeds into $\mathbf{P}'$ while $G$ no longer feeds into $\mathbf{P}'$ (see Figures 3 and 4 for an example). The idea is to create a sufficient number of clones of $G$ so that $G$ and its clones become part of some cones and hence can be abstracted away. Repeated applications of this operation can allow an otherwise unmanageable system to have a small enough abstraction for compilation and diagnosis to succeed. The hierarchical algorithm is then extended to diagnose the new system and the result mapped to the original system. We show that we can now solve almost all the benchmark circuits, using this approach.

Before we go into the details of the new method, we differentiate it from a technique known as *node splitting* (Choi, Chavira, & Darwiche, 2007), which is used to solve MPE queries on a Bayesian network. Node splitting breaks enough number of edges between nodes from the network such that the MPE query on the resulting network becomes easy to solve. A broken edge is replaced with a root variable with a uniform prior. The resulting network is a relaxation or approximation of the original in that its MPE solution, which may be computed from its compilation, gives an upper bound on the MPE solution of the original network. A depth-first branch and bound search algorithm then searches for an optimal solution using these bounds to prune its search space. A similar approach is also used to solve Weighted Max-SAT problems (Pipatsrisawat & Darwiche, 2007).

This version of node splitting is not directly applicable in the present setting for the following reasons. If edges in a system are broken and redirected into new root variables (primary inputs), the resulting system represents a different input-output function from that of the original system. The abnormal observation on the original system may hence become a normal one on the new system (if the edges through which the fault propagates are broken), eliminating the basis for diagnosis. Our technique of component cloning, which can also be viewed as a version of node splitting, introduces clones of a component instead of primary inputs and preserves the input-output function of the system. Also, the new system is a relaxation of the original in that its diagnoses are a superset of those of the original.

We now formally define component cloning:

**Definition 2** (Component Cloning). *Let $G$ be a component in a system $\mathbf{C}$ with parents $\mathbf{P}$. We say that $G$ is **cloned according to parents** $\mathbf{P}' \subset \mathbf{P}$ when system $\mathbf{C}$ results in a system $\mathbf{C}'$ as follows:*

- *The edges going from $G$ to its parents $\mathbf{P}'$ are removed.*

- *A new component $G'$ functionally equivalent to $G$ is added to the system such that $G'$ shares the inputs of $G$ and feeds into each of $\mathbf{P}'$.*

Figures 3 and 4 show an example where creating a clone $B'$ of $B$ according to $\{D\}$ results in a new circuit whose abstraction contains only the gates $\{A, D, K, V\}$, whereas the abstraction of the original circuit contains also gate $B$.

## 5.1 Choices in Component Cloning

There are two choices to be made in component cloning: Which components do we clone, and for each of them how many clones do we create and how do they split the parents?

Since the goal of cloning is to reduce the abstraction size, it is clear that we only wish to clone those components that lie in the abstraction (i.e., not within cones). Among these, cloning of the root of a cone cannot reduce the abstraction size as it will destroy the existing cone by reintroducing some of the components inside the cone into the abstraction. For example, cloning $D$ according to $K$ in Figure 4 will produce a circuit where $D$ and its clone can be abstracted away but $B'$ is no longer dominated by $D$ and hence is reintroduced into the abstraction. Therefore, the final candidates for cloning are precisely those components in the abstract system that are not roots of cones. Note that the order in which these candidates are processed is unimportant in that each when cloned will produce an equal reduction, namely a reduction of precisely 1 in the abstraction size, if any.

It then remains to determine for each candidate how many clones to create and how to connect them to the parents. To understand our final method, it helps to consider a naive method that simply creates $|\mathbf{P}| - 1$ clones (where $\mathbf{P}$ is the set of parents) and has each clone, as well as the original, feed into exactly one parent. This way every parent of the component becomes the root of a cone and the component itself and all its clones are abstracted away. In Figure 3, for example, $B$ has three parents $\{E, A, D\}$, and this naive method would create two clones of $B$ for a total of three instances of the gate to split the three parents, which would result in the same abstraction as in Figure 4.

The trick now is that the number of clones can be reduced by knowing that some parents of the component may lie in the same cone and a single clone of the component according to those parents will be sufficient for that clone to be abstracted away. In the example of Figure 3, again, the parents $E, A$ of $B$ lie in the same cone $A$ and it would suffice to create a single clone of $B$ according to $\{E, A\}$, resulting in the same, more efficient cloning as in Figure 4.

More formally, we partition the parents of a component $G$ into subsets $\mathbf{P_1}, \mathbf{P_2}, \ldots, \mathbf{P_q}$ such that those parents of $G$ that lie in the same cone are placed in the same subset and the rest in separate ones. We then create $q - 1$ clones of $G$ according to any $q - 1$ of these subsets, resulting in $G$ and all its clones being abstracted away. This process is repeated for each candidate component until the abstraction size is small enough or no further reduction is possible.

## 5.2 Diagnosis with Component Cloning

The new system is functionally equivalent to the original and has a smaller abstraction, but is not equivalent to the original for diagnostic purposes. As the new model allows a component and its clones to fail independently of each other, it is a relaxation of the original model in that the diagnoses of the new system form a superset of those of the original. Specifically, each diagnosis of the new system that assigns the same health state to a component and its clones for all components corresponds to a diagnosis of the original system; other diagnoses are spurious and are to be ignored.

The core diagnosis process given in Algorithm 2 continues to be applicable on the new system, with only two minor modifications necessary. First, the spurious diagnoses are (implicitly) filtered out by assuming the same health state for all clones (including the original) of a component as soon as the health state of any one of them is known. Second, whenever measurement of a clone of a component is proposed, the actual measurement is taken on the original component in the original system, for obvious reasons (in other words, the new system is used for reasoning and the original for measurements).

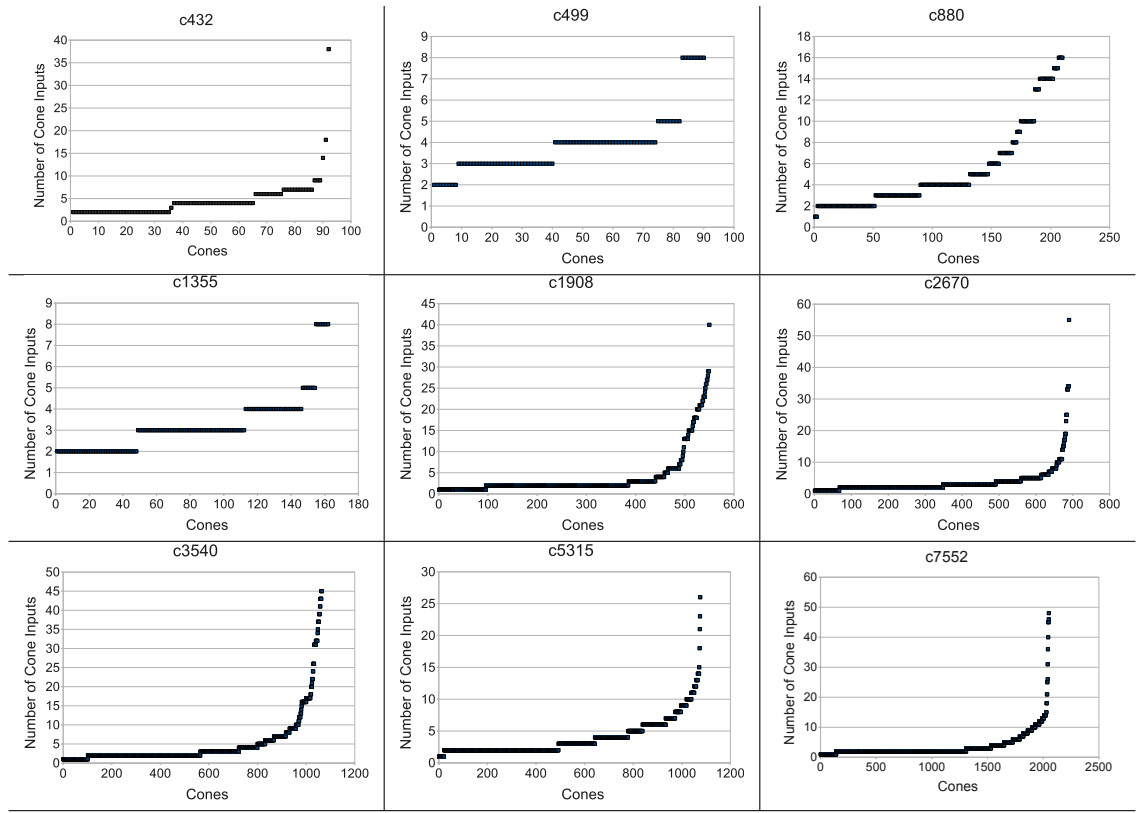

Figure 5: Cones in ISCAS-85 circuits.

In principle, the presence of spurious diagnoses in the model can potentially skew the measurement point selection heuristic (at least in the early stages of diagnosis, before the spurious diagnoses are gradually filtered out). However, by using smaller benchmarks that could be diagnosed both with and without cloning, we conducted an empirical analysis which indicates, interestingly, that the overall diagnostic cost is only slightly affected. We discuss this in more detail in Section 7.3.

## 6. Diagnostic Cost Estimation

We now address an interesting issue stemming from an observation we made conducting experiments (to be detailed in the next section): While system abstraction is always beneficial to compilation, the diagnostic cost does not always improve with the associated hierarchical diagnosis. On the one hand, the hierarchical diagnosis approach can help in cases which otherwise result in high costs using baseline approach by quickly finding faulty portions of the system, represented by a set of faulty cones, and then directing the sequential diagnosis to take measurements inside those cones, resulting in more useful measurements. On the other hand, it can introduce overhead for cases where it has to needlessly go through hier-

345

archies to locate the actual faults, and measure inputs of cones involved, while the baseline version can find them more directly and efficiently.

The overhead of hierarchical approach can be quite high for faults that lie in cones with a large number of inputs. For example, the graphs in Figure 5 show the number of inputs, represented as dots, of various cones in ISCAS-85 circuits. Note that most of the cones have a small number of inputs; however, some cones can have more than 30 inputs, especially in $c432$ and the circuits beyond $c1908$, which contribute to increased diagnostic cost in several cases (such increase in the cost due to cones was also confirmed by a separate set of experiments using a large set of systematically generated combinational circuits, detailed in Appendix C). To avoid the potential high cost of diagnosis for faults that lie in a cone with a large number of inputs it is tempting to destroy that cone before compilation so that any fault in it can now be directly found. However, due to the associated increase in the abstraction size, destroying cones may cause increased costs for those cases that could previously be solved more efficiently, and thus may show a negative impact, overall. This calls for an automatic mechanism to predict the effect of destroying certain cones on the overall diagnostic cost, which is the subject of this section.

We propose a novel cost estimation function to predict the average diagnostic cost when a given abstraction of the system is considered for diagnosis, where different abstractions can be obtained by destroying different cones in the system. Since cones can be destroyed automatically, the function can be used to automatically propose an abstraction of the system, to be used for diagnosis, that is more likely to give optimal average cost. The function uses only the hierarchical structure of the given abstraction to predict its cost and does not take into account other parameters that may also contribute to the cost, such as the probabilities. In addition the function is limited to single fault cases only. Therefore, the expected cost computed by this function is only indicative and cannot be always correct. However, experiments show that the function is often quite useful in proposing an abstraction of the system that is more likely to give optimal cost (to be discussed in the next section).

To estimate the expected diagnostic cost we assume that it is composed of two quantities namely the *isolation cost* and the *abstraction cost*, which are inversely proportional to each other. The *isolation cost* captures how well the given system abstraction can isolate the faulty portions of the system. Therefore the isolation cost is minimum when a complete abstraction of the system is used (i.e., all cones are considered) and generally increases as cones are destroyed. The *abstraction cost* captures the overhead cost due to introduction of cones. Hence, the abstraction cost is minimum (zero) when no abstraction is considered and generally increases as cones are introduced.

We define the isolation cost of diagnosis considering an abstraction of the system to be the average cost required to isolate a single fault in the system using that abstraction. Similarly, we define the abstraction cost of diagnosis to be the average overhead cost required to diagnose a single fault in the system using that abstraction. Then the expected average cost of diagnosis when an abstraction of the system is considered for diagnosis is the sum of the isolation and the abstraction costs for that abstraction. As different cones are destroyed in a given abstraction of the system we expect changes in the values of the abstraction and isolation costs, which determine whether the overall cost can go up or down (if the changes are uneven) or stay constant (if the changes are even). The idea is to obtain an abstraction of the system to strike a balance between the two quantities to get an overall optimal cost. Below we discuss how the isolation and abstraction costs can be estimated.

We noted in our experiments when using the baseline approach that our heuristic can isolate a single fault in the system with a cost that is on average comparable to the $log_2$ of the number of measurement points in the system, which provided us with the basis for computing the isolation cost. In the hierarchical approach, when a fault lies inside a cone one can first estimate the isolation cost of diagnosing the cone, separately, and then add it to the isolation cost of diagnosing the abstract system to get the average isolation cost for all (single) faults that lie in that cone. For example, when no cones are considered the cost of isolating a fault in the circuit in Figure 3 is $log_2(6) = 2.58$ (values of $P$, $Q$, $R$ and $V$ are already known). However, when cones are considered the cost of isolating a fault that lies inside the cone $A$ is the sum of the isolation cost of the abstract circuit and the isolation cost of the subcircuit inside cone $A$, which is $log_2(4) + log_2(1) = 2$. Similarly, to get an average isolation cost for all single faults in the system, when using the hierarchical approach, one can add the isolation cost of diagnosing the abstract system and the average of the isolation costs of diagnosing all the abstract components (where the isolation cost for an abstract component which is not a cone is zero). Note that the isolation cost of diagnosing a cone can be computed by again taking the abstraction of the cone.

To estimate the abstraction cost of diagnosis under a given abstraction we first need to estimate the overhead cost involved for each individual component in the system under that abstraction. To estimate the overhead cost of a, possibly faulty, component one can take the union of all the inputs and outputs of cones in which that component lies, and the number of such measurement points (approximately) constitutes the required overhead cost for that component. If a component does not lie in any cone then the overhead cost for that component is zero. For example, when the circuit in Figure 3 is diagnosed using the hierarchical approach, to find the gate $J$ as faulty one must first find the cone $A$ to be faulty and then the cone $E$ to be faulty and then the gate $J$ to be faulty. So the overhead cost for the gate $J$ in this case will be $1 + 2 + 1 = 4$ (i.e., we have to measure wires $A$, $B$, $E$, $J$, assuming that $Q$ is known). The abstraction cost of diagnosis under a given abstraction of the system is then the average of the overhead costs of all the system components under that abstraction.

We now give formal definitions related to the cost estimation function. Let $MP_u(\mathbf{C})$ be the set of those measurement points in the system $\mathbf{C}$ whose values are unknown, and $MP_u(G)$ the set of those inputs and output of an abstract or concrete component $G$ whose values are unknown. Let $p$ be the number of abstract components in an abstraction $\mathbf{A_C}$ of system $\mathbf{C}$. Let $G_i \in \mathbf{A_C}$ be an abstract component (either a concrete component or a cone in the abstraction; a concrete component in the abstraction can be regarded as a trivial cone containing only the component itself). Let $\mathbf{D_{G_i}}$ be the subsystem dominated by $G_i$ and $\mathbf{A_{G_i}}$ be the abstraction of the subsystem.

The isolation cost $IC(\mathbf{C}, \mathbf{A_C})$ when an abstraction $\mathbf{A_C}$ of the system $\mathbf{C}$ is considered for diagnosis is the sum of $log_2(|MP_u(\mathbf{A_C})|)$ and the average of the isolation costs computed, in a similar manner, for the subsystems contained in the abstract components in $\mathbf{A_C}$:

$$IC(\mathbf{C}, \mathbf{A_C}) = \begin{cases} log_2(|MP_u(\mathbf{A_C})|) + \frac{1}{p} \sum_{i=1}^{p} IC(\mathbf{D_{G_i}}, \mathbf{A_{G_i}}), & \text{if } |MP_u(\mathbf{A_C})| > 0 \\ \frac{1}{p} \sum_{i=1}^{p} IC(\mathbf{D_{G_i}}, \mathbf{A_{G_i}}) & \text{otherwise} \end{cases} \quad (5)$$

where $IC(\mathbf{D_{G_i}}, \mathbf{A_{G_i}})$ recursively computes the isolation cost of the subsystem contained in the abstract component $G_i$, using Equation 5, by taking its abstraction $\mathbf{A_{G_i}}$. Note that when computing $IC(\mathbf{D_{G_i}}, \mathbf{A_{G_i}})$ we assume that the inputs and output of $G_i$ have already been measured. Thus $MP_u(\mathbf{D_{G_i}})$ excludes the inputs and output of cone $G_i$. If $G_i$ is a concrete component then $IC(\mathbf{D_{G_i}}, \mathbf{A_{G_i}}) = 0$. If no cones are considered ($\mathbf{A_C} = \mathbf{C}$) then $\sum_{i=1}^{p} IC(\mathbf{D_{G_i}}, \mathbf{A_{G_i}}) = 0$ and the isolation cost is simply equal to $log_2(|MP_u(\mathbf{C})|)$).

To compute the abstraction cost of diagnosing the system under a given abstraction we first compute the overhead costs of diagnosing individual cones in the abstraction. Then we multiply the abstraction cost for a cone with the number of components contained in that cone to get the total overhead cost for all the components in that cone. Adding up the overhead costs computed this way from all the cones in the abstraction and dividing this number by the total number of concrete components in the whole system gives us the average overhead cost per component, which we call the abstraction cost. Formally: Let there be $q$ cones in $\mathbf{A_C}$. Then the abstraction cost $AC(\mathbf{C}, \mathbf{A_C})$ when the abstraction $\mathbf{A_C}$ of the system $\mathbf{C}$ is considered for diagnosis is given as:

$$AC(\mathbf{C}, \mathbf{A_C}) = \frac{1}{n} \sum_{i=1}^{q} |\mathbf{D_{G_i}}| * \{MP_u(G_i) + AC(\mathbf{D_{G_i}}, \mathbf{A_{G_i}})\} : G_i \in \mathbf{A_C} \text{ is a cone} \quad (6)$$

where $|\mathbf{D_{G_i}}|$ is the number of (concrete) components contained in the cone $G_i$, and $MP_u(G_i) + AC(\mathbf{D_{G_i}}, \mathbf{A_{G_i}})$ recursively computes the abstraction cost of diagnosing the cone $G_i$, using Equation 6, by taking its abstraction $\mathbf{A_{G_i}}$. When the abstraction cost of $G_i$ is multiplied by $|\mathbf{D_{G_i}}|$ we effectively add the cost of measuring cone inputs and output in the overhead cost of every component inside the cone. Again note that when computing $AC(\mathbf{D_{G_i}}, \mathbf{A_{G_i}})$ we assume that all the variables in $MP_u(G_i)$ have already been measured. Thus $MP_u(\mathbf{D_{G_i}})$ excludes the inputs and output of cone $G_i$.

Finally the total expected cost $EDC(\mathbf{C}, \mathbf{A_C})$ of diagnosing a system $\mathbf{C}$ when an abstraction $\mathbf{A_C}$ of the system is considered for diagnosis is given as:

$$EDC(\mathbf{C}, \mathbf{A_C}) = IC(\mathbf{C}, \mathbf{A_C}) + AC(\mathbf{C}, \mathbf{A_C}). \quad (7)$$

## 7. Experimental Results

This section provides an empirical evaluation of our new diagnostic system, referred to as SDA (sequential diagnosis by abstraction), that implements the baseline, hierarchical, and cloning-based approaches described in Sections 4 and 5, and the cost estimation function described in Section 6. All experiments were conducted on a cluster of 32 computers consisting of two types of (comparable) CPUs, Intel Core Duo 2.4 GHz and AMD Athlon 64 X2 Dual Core Processor 4600+, both with 4 GB of RAM running Linux. A time limit of 2 hours and a memory limit of 1.5 GB were imposed on each test case. The d-DNNF compilation was done using the publicly available d-DNNF compiler c2d (Darwiche, 2004, 2005). The CNF was simplified before compilation using the given observation, which allowed us to compile more circuits, at the expense of requiring a fresh compilation per observation (see Algorithm 2, line 1).

We generated single- and multiple-fault scenarios using ISCAS-85 benchmark circuits, where in each scenario a set of gates is assumed to be faulty. For single-fault cases of circuits up to $c1355$ we simulated the equal prior probability of faults by generating $n$ fault scenarios for each circuit, where $n$ equals the number of gates in the circuit: Each scenario contains a different faulty gate. We then randomly generated 5 test cases (abnormal observations) for each of these $n$ scenarios. Doing the same for multiple-fault scenarios would not be practical due to the large number of combinations, so for each circuit up to $c1355$ (respectively, larger than $c1355$) we simply generated 500 (respectively, 100) random scenarios with the given fault cardinality and a random test case for each scenario.

Thus in each test case we have a faulty circuit where some gate or gates give incorrect outputs. The inputs and outputs of the circuit are observed. The values of internal wires are then computed by propagating the inputs in the normal circuit towards the outputs followed by propagating the outputs of the assumed faulty gates one by one such that deeper faults are propagated first. The obtained values of internal wires are then used to simulate the results of taking measurements. We use $Pr(okX = 1) = 0.9$ for all gates $X$ of the circuit. Note that such cases, where all gates fail with equal probability, are conceivably harder to solve as the diagnoses will tend to be less differentiable. Then, for each gate, the two output values are given equal probability when the gate is faulty. Again, this will tend to make the cases harder to solve due to the high degree of uncertainty. For each circuit and fault cardinality, we report the cost (number of measurements taken) and time (including the compilation time, in CPU seconds) to locate the faults, averaged over all test cases solved.

We present the experiments in four subsections demonstrating the effectiveness of the four techniques proposed in this paper, namely the new heuristic, hierarchical sequential diagnosis, component cloning, and the cost estimation function.

## 7.1 Effectiveness of Heuristic

We start with a comparison of the baseline algorithm of SDA with GDE and show that SDA achieves similar diagnostic costs and scales to much larger circuits, hence illustrating the effectiveness of our new heuristic (along with the new way to compute probabilities).

### 7.1.1 Comparison with GDE

We could obtain only the tutorial version of GDE (Forbus & de Kleer, 1993) for the comparison, downloadable from http://www.qrg.northwestern.edu/BPS/readme.html. GDE uses ATCON, a constraint language developed using the LISP programming language, to represent diagnostic problem cases. A detailed account of this language is given by Forbus and de Kleer (1993). Further, it employs an interactive user interface that proposes measurement points with their respective costs and lets the user enter outcomes of measurements. For the purpose of comparison we translated our problem descriptions to the language accepted by GDE, and also modified GDE to automatically read in the measurement outcomes from the input problem description. We also compiled the LISP code to machine dependent binary code using the native C compiler to improve run-time performance.

| size | system | single-fault | | double-fault | | triple-fault | |
|------|--------|------|------|------|------|------|------|
| | | *cost* | *time* | *cost* | *time* | *cost* | *time* |
| **13** | GDE | 3.6 | 2.0 | 3.8 | 1.81 | 4.0 | 1.9 |
| | SDA | 3.6 | 0.01 | 3.4 | 0.01 | 2.8 | 0.01 |
| **14** | GDE | 3.5 | 6.66 | 3.3 | 15.1 | 3.0 | 14 |
| | SDA | 4.2 | 0.01 | 2.9 | 0.01 | 2.9 | 0.01 |
| **15** | GDE | 3.4 | 111 | 3.5 | 88 | 4.3 | 299 |
| | SDA | 3.9 | 0.01 | 3.4 | 0.01 | 3.7 | 0.01 |
| **16** | GDE | 3.3 | 398 | 3.5 | 556 | 3.2 | 509 |
| | SDA | 3.5 | 0.01 | 3.3 | 0.01 | 2.8 | 0.01 |
| **17** | GDE | 3.7 | 2876 | 4.6 | 4103 | 4.5 | 2067 |
| | SDA | 3.8 | 0.01 | 4.2 | 0.01 | 4.2 | 0.01 |

Table 1: Comparison with GDE.

This version of GDE, developed for tutorial purposes, computes the set of *minimal diagnoses* instead of probable diagnoses. This makes our comparison less informative. Nevertheless, we are able to make a reasonable comparison in terms of diagnostic cost as the set of minimal diagnoses can also serve as a large set of probable diagnoses when components have equal prior probabilities. According to de Kleer (1992) availability of more diagnoses aids in heuristic accuracy, whereas focusing on a smaller set of probable diagnoses can be computationally more efficient but increase the average diagnostic cost.

This version of GDE was in fact unable to solve any circuit in ISCAS-85. To enable a useful comparison, we extracted a set of small subcircuits from the ISCAS-85 circuits: 50 circuits of size 13, 14, 15 and 16, and 10 circuits of size 17. For each circuit we randomly generated 5 single-fault, 5 double-fault, and 5 triple-fault scenarios, and one test case (input/output vector) for each fault scenario. The comparison between GDE and SDA (baseline) on these benchmarks given in Table 1 shows that SDA performs as well as GDE in terms of diagnostic cost.

### 7.1.2 Larger Benchmarks

To evaluate the performance of SDA on the larger ISCAS-85 circuits, we have again conducted three sets of experiments, this time involving single, double, and five faults, respectively. As the version of GDE available to us is unable to handle these circuits, in order to provide a systematic reference point for comparison we have implemented a random strategy where a random order of measurement points is generated for each circuit and used for all the test cases. This strategy also uses the d-DNNF to check whether the stopping criteria have been met.

Table 2 shows the comparison between the random strategy and SDA using the baseline approach with two different heuristics, one based on entropies of wires alone (ew) and the other based also on failure probabilities (fp). For each of the three systems we ran the same set of experiments with and without pruning the d-DNNF (using the known fault cardinality as described in Section 4.1.2), indicated in the third column of the table. Only the test cases for the first four circuits could be solved. For other circuits the failure occurred during the compilation phase, and hence affected both the random strategy and SDA.

| circuit | system | pruning | single-fault | | double-fault | | five-fault | |
|---------|--------|---------|------|------|------|------|------|------|
| | | | *cost* | *time* | *cost* | *time* | *cost* | *time* |
| **c432** | RAND | *no* | 92.3 | 20.7 | 97.7 | 23.2 | 117.8 | 26.5 |
| | | *yes* | 4.5 | 11.4 | 36.8 | 12.4 | 99.7 | 17.2 |
| (160 gates) | SDA(ew) | *no* | 42.0 | 16.6 | 42.5 | 21.3 | 68.4 | 25.5 |
| | | *yes* | 3.7 | 11.1 | 8.6 | 12.0 | 33.8 | 12.8 |
| | SDA(fp) | *no* | 6.7 | 11.7 | 6.4 | 12.5 | 9.4 | 13.0 |
| | | *yes* | 4.3 | 11.0 | 5.0 | 12.3 | 9.1 | 12.6 |
| **c499** | RAND | *no* | 109.6 | 0.8 | 120.6 | 1.2 | 150.0 | 1.4 |
| | | *yes* | 5.5 | 0.2 | 20.1 | 0.2 | 104.9 | 0.7 |
| (202 gates) | SDA(ew) | *no* | 58.1 | 0.7 | 54.0 | 0.5 | 95.8 | 0.8 |
| | | *yes* | 3.6 | 0.2 | 3.7 | 0.2 | 35.7 | 0.3 |
| | SDA(fp) | *no* | 6.5 | 0.2 | 4.3 | 0.2 | 7.2 | 0.2 |
| | | *yes* | 4.8 | 0.2 | 3.0 | 0.2 | 7.1 | 0.2 |
| **c880** | RAND | *no* | 221.0 | 1.9 | 251.3 | 1.9 | 306.4 | 2.3 |
| | | *yes* | 5.4 | 0.2 | 47.3 | 0.3 | 205.7 | 1.3 |
| (383 gates) | SDA(ew) | *no* | 26.8 | 0.3 | 32.8 | 0.4 | 79.0 | 0.7 |
| | | *yes* | 4.0 | 0.2 | 6.8 | 0.2 | 30.5 | 0.4 |
| | SDA(fp) | *no* | 10.8 | 0.2 | 9.2 | 0.2 | 15.8 | 0.3 |
| | | *yes* | 5.6 | 0.2 | 6.7 | 0.2 | 14.0 | 0.3 |
| **c1355** | RAND | *no* | 327.2 | 4.3 | 365.7 | 5.7 | 437.4 | 5.6 |
| | | *yes* | 7.4 | 0.4 | 59.0 | 1.0 | 328.6 | 3.5 |
| (546 gates) | SDA(ew) | *no* | 82.6 | 1.3 | 91.2 | 1.5 | 203.9 | 3.4 |
| | | *yes* | 4.9 | 0.4 | 5.5 | 0.4 | 65.9 | 1.1 |
| | SDA(fp) | *no* | 34.1 | 0.8 | 14.8 | 0.5 | 19.3 | 0.8 |
| | | *yes* | 8.0 | 0.4 | 9.4 | 0.6 | 18.4 | 0.6 |

Table 2: Effectiveness of heuristic.

It is clear that the diagnostic cost is significantly lower with both heuristics of SDA than with the random strategy whether or not pruning has been used. It is also interesting to note that pruning significantly reduces the diagnostic cost for the random and SDA-ew strategies, but has much less effect on SDA-fp except in a few cases (c1355 single-fault). Moreover, SDA-fp generally dominates SDA-ew, both with and without pruning.

We may also observe that (i) on the five-fault cases, SDA-fp without pruning results in much lower diagnostic cost than SDA-ew with pruning; (ii) on the double-fault cases, the two are largely comparable; and (iii) on the single-faults cases, the comparison is reversed. This indicates that as the fault cardinality rises, the combination of failure probabilities and wire entropies appears to achieve an effect similar to that of pruning. That SDA-ew with pruning performs better than SDA-fp without pruning on single-fault cases can be attributed to the fact that on these cases pruning is always exact and hence likely to result in maximum benefit.

## 7.2 Effectiveness of Abstraction

We now report, in Table 3, the results of repeating the same experiments with SDA-fp using the hierarchical approach.

Most notably, the running time generally reduces for all cases and we are now able to handle two more circuits, namely *c*1908 and *c*2670, solving 139 of 300 cases for *c*1908 (25 of single-, 15 of double-, and 99 of five-fault cases) and 258 of 300 cases for *c*2670 (100 of

| circuit | pruning | single-fault | | double-fault | | five-fault | |
|---|---|---|---|---|---|---|---|
| | | *cost* | *time* | *cost* | *time* | *cost* | *time* |
| **c432** | *no* | 15.4 | 0.4 | 15.8 | 0.5 | 22.2 | 0.5 |
| (64 cones) | *yes* | 4.9 | 0.3 | 10.4 | 0.4 | 21.5 | 0.4 |
| **c499** | *no* | 7.3 | 0.1 | 5.8 | 0.1 | 10.5 | 0.2 |
| (90 cones) | *yes* | 4.5 | 0.1 | 3.9 | 0.1 | 9.6 | 0.2 |
| **c880** | *no* | 9.5 | 0.1 | 10.2 | 0.1 | 17.4 | 0.2 |
| (177 cones) | *yes* | 5.6 | 0.1 | 7.6 | 0.1 | 16.3 | 0.2 |
| **c1355** | *no* | 9.3 | 0.3 | 8.2 | 0.2 | 14.0 | 0.3 |
| (162 cones) | *yes* | 5.8 | 0.2 | 6.3 | 0.2 | 14.4 | 0.3 |
| **c1908** | *no* | 11.0 | 222 | 17.1 | 587 | 34.9 | 505 |
| (374 cones) | *yes* | 3.0 | 214 | 8.5 | 463 | 32.4 | 383 |
| **c2670** | *no* | 16.3 | 213 | 19.2 | 172 | 25.4 | 58 |
| (580 cones) | *yes* | 6.5 | 196 | 13.3 | 90 | 24.3 | 45 |

Table 3: Effectiveness of abstraction.

| circuit | total gates | abstraction size | cloning time | total clones | abstraction size after cloning |
|---|---|---|---|---|---|
| **c432** | 160 | 59 | 0.03 | 27 | 39 |
| **c499** | 202 | 58 | 0.02 | 0 | 58 |
| **c880** | 383 | 77 | 0.1 | 24 | 57 |
| **c1355** | 58 | 58 | 0.05 | 0 | 58 |
| **c1908** | 880 | 160 | 0.74 | 237 | 70 |
| **c2670** | 1193 | 167 | 0.77 | 110 | 116 |
| **c3540** | 1669 | 353 | 5.64 | 489 | 165 |
| **c5315** | 2307 | 385 | 3.6 | 358 | 266 |
| **c6288** | 2416 | 1456 | 0.16 | 0 | 1456 |
| **c7552** | 3512 | 545 | 6.68 | 562 | 378 |

Table 4: Results of preprocessing step of cloning.

single-, 60 of double-, and 98 of five-fault cases). Again all failures occurred during the compilation phase. Note that some observations do not cause sufficient simplification of the theory for it to be successfully compiled even after abstraction. In terms of diagnostic cost, in most cases the hierarchical approach is comparable to the baseline approach. On *c*432, the baseline approach consistently performs better than the hierarchical in each fault cardinality, while the reverse is true on *c*1355. Note also that pruning helps further reduce the diagnostic cost to various degrees as with the baseline approach.

As discussed earlier, the results confirm that the main advantage of hierarchical approach is that larger circuits can be solved. For circuits that can also be solved by the baseline approach, hierarchical approach may help reduce the diagnostic cost by quickly finding faulty portions of the circuit, represented by a set of faulty cones, and then directing the measurements inside them, which can result in more useful measurements (e.g. in the case of *c*1355). On the other hand, it may suffer in cases where it has to needlessly go through hierarchies to locate the actual faults, while the baseline version can find them more directly and efficiently (e.g. in the case of *c*432). This is further discussed in Section 7.4.

| circuit | single-fault | | double-fault | | five-fault | |
|---|---|---|---|---|---|---|
| | *cost* | *time* | *cost* | *time* | *cost* | *time* |
| **c432** | 7.2 | 10.3 | 6.6 | 7.8 | 9.6 | 9.7 |
| **c880** | 11.2 | 0.2 | 9.3 | 0.2 | 16.2 | 0.3 |

Table 5: Effect of component cloning on diagnostic performance.

| circuit | single-fault | | double-fault | | five-fault | |
|---|---|---|---|---|---|---|
| | *cost* | *time* | *cost* | *time* | *cost* | *time* |
| **c432** | 15.2 | 0.1 | 14.8 | 0.1 | 20.2 | 0.1 |
| **c880** | 8.8 | 0.1 | 9.3 | 0.1 | 15.8 | 0.2 |
| **c1908** | 13.6 | 2.8 | 18.3 | 5.0 | 35.4 | 5.1 |
| **c2670** | 13.5 | 4.5 | 15.3 | 0.7 | 20.1 | 2.3 |
| **c3540** | 27.8 | 382 | 30.5 | 72.5 | 36.1 | 108.6 |
| **c5315** | 7.2 | 2.5 | 21.1 | 5.9 | 24.4 | 6.6 |
| **c7552** | 70.6 | 1056 | 43.1 | 129.0 | 104.8 | 1108 |

Table 6: Hierarchical sequential diagnosis with component cloning ($c$499 and $c$1355 omitted as they are already easy to diagnose and cloning does not lead to reduced abstraction).

## 7.3 Effectiveness of Component Cloning

In this subsection we discuss the experiments with component cloning. We show that cloning does not significantly affect diagnostic cost and allows us to solve much larger circuits, in particular, nearly all the circuits in the ISCAS-85 suite.

Table 4 shows the result of the pre-processing step of cloning on each circuit. The columns give the name of the circuit, the total number of gates in that circuit, the size of the abstraction of the circuit before cloning, the time spent on cloning, the total number of clones created in the circuit, and the abstraction size of the circuit obtained after cloning. On all circuits except $c$499, $c$1355, and $c$6288, a significant reduction in the abstraction size has been achieved. $c$6288 appears to be an extreme case with a very large abstraction that lacks hierarchy; while gates in the abstractions of $c$499 and $c$1355 are all roots of cones, affording no opportunities for further reduction (note that these two circuits are already very simple and easy to diagnose).

We start by investigating the effect of component cloning on diagnostic performance. To isolate the effect of component cloning we use the baseline version of SDA (i.e., without abstraction), and without pruning. Table 5 summarizes the performance of baseline SDA with cloning on the circuits $c$432 and $c$880. Comparing these results with the corresponding entries in Table 2 shows that the overall diagnostic cost is only slightly affected by cloning. We further observed that in a significant number of cases the proposed measurement sequence did not change after cloning, while in most of the other cases it changed only insubstantially. Moreover, in a number of cases, although a substantially different sequence of measurements was proposed, the actual diagnostic cost did not change much. Finally, note that the diagnosis time in the case of $c$432 has reduced after cloning, which can be ascribed to the general reduction in the complexity of compilation due to a smaller abstraction.

| circuit | total cases | max. cone inputs | abstraction size | measurement points | AC | IC | EDC | cases solved | single-fault | |
|---|---|---|---|---|---|---|---|---|---|---|
| | | | | | | | | | *cost* | *time* |
| **c432** | 800 | 38 | 39 | 32 | 11.51 | 5.67 | 17.1 | 800 | 15.2 | 0.06 |
| | | 18 | 49 | 42 | 5.22 | 6.05 | 11.2 | 800 | 11.0 | 0.1 |
| | | 14 | 52 | 45 | 4.87 | 6.11 | 10.9 | 800 | 11.0 | 0.1 |
| | | 9 | 53 | 46 | 4.64 | 6.14 | 10.8 | 800 | 10.7 | 0.1 |
| | | 4 | 104 | 97 | 2.11 | 6.72 | 8.8 | 800 | 8.8 | 0.3 |
| | | 0 | 187 | 180 | 0.00 | 7.50 | 7.5 | 800 | 7.3 | 7.3 |
| **c499** | 1010 | 8 | 58 | 26 | 3.77 | 5.32 | 9.0 | 1010 | 7.3 | 0.1 |
| | | 5 | 74 | 42 | 3.13 | 5.91 | 9.0 | 1010 | 7.7 | 0.1 |
| | | 3 | 170 | 138 | 0.71 | 7.10 | 7.8 | 1010 | 9.4 | 0.1 |
| | | 0 | 202 | 170 | 0.0 | 7.40 | 7.4 | 1010 | 6.4 | 0.1 |
| **c880** | 1915 | 16 | 57 | 31 | 6.54 | 5.42 | 11.9 | 1915 | 8.7 | 0.1 |
| | | 14 | 74 | 48 | 5.75 | 6.02 | 11.7 | 1915 | 8.5 | 0.1 |
| | | 10 | 105 | 79 | 4.22 | 6.72 | 10.9 | 1915 | 8.0 | 0.1 |
| | | 6 | 170 | 144 | 2.70 | 7.48 | 10.1 | 1915 | 8.6 | 0.1 |
| | | 0 | 407 | 381 | 0.0 | 8.57 | 8.5 | 1915 | 10.8 | 0.2 |
| **c1355** | 2730 | 8 | 58 | 26 | 3.59 | 6.34 | 9.9 | 2730 | 9.30 | 0.1 |
| | | 5 | 98 | 66 | 2.74 | 7.20 | 9.9 | 2730 | 12.55 | 0.2 |
| | | 4 | 114 | 82 | 2.47 | 7.27 | 9.7 | 2730 | 12.39 | 0.2 |
| | | 3 | 266 | 234 | 1.43 | 8.23 | 9.6 | 2730 | 22.5 | 0.3 |
| | | 2 | 426 | 394 | 0.43 | 8.77 | 9.2 | 2730 | 33.5 | 0.4 |
| | | 0 | 546 | 514 | 0.0 | 9.00 | 9.0 | 2730 | 34.0 | 0.4 |
| **c1908** | 859 | 40 | 70 | 45 | 14.37 | 7.07 | 21.4 | 859 | 18.7 | 2.6 |
| | | 29 | 76 | 51 | 12.85 | 7.15 | 20.0 | 859 | 17.8 | 5.8 |
| | | 28 | 80 | 55 | 12.70 | 7.23 | 19.9 | 859 | 18.3 | 5.9 |
| | | 27 | 82 | 57 | 12.62 | 7.27 | 19.8 | 859 | 18.2 | 5.9 |
| | | 20 | 138 | 113 | 8.36 | 7.82 | 16.2 | 859 | 17.7 | 15.0 |
| | | 18 | 150 | 125 | 7.79 | 7.92 | 15.7 | 859 | 17.7 | 47.5 |
| **c2670** | 989 | 55 | 56 | 52 | 17.84 | 6.40 | 24.2 | 989 | 19.2 | 0.7 |
| | | 34 | 58 | 57 | 16.19 | 6.53 | 22.7 | 989 | 19.1 | 0.8 |
| | | 33 | 128 | 64 | 15.63 | 6.68 | 22.3 | 989 | 18.6 | 0.8 |
| | | 25 | 178 | 114 | 11.52 | 7.44 | 18.9 | 970 | 16.1 | 79.0 |

Table 7: Effectiveness of diagnostic cost estimation.

Our final set of experimental results with ISCAS-85 circuits, summarized in Table 6, illustrates the performance of hierarchical sequential diagnosis with component cloning—the most scalable version of SDA. All the test cases for circuits $c1908$ and $2670$ were now solved, and the largest circuits in the benchmark suite could now be handled: All the cases for $c5315$, 164 of the 300 cases for $c3540$ (34 of single-, 65 of double-, and 65 of five-fault cases), and 157 of the 300 cases for $c7552$ (60 of single-, 26 of double-, and 71 of five-fault cases) were solved. In terms of diagnostic cost cloning generally resulted in a slight improvement. In terms of time the difference is insignificant for $c432$ and $c880$, and for the larger circuits ($c1908$ and $c2670$) diagnosis with cloning was clearly more than an order of magnitude faster.

## 7.4 Effectiveness of Diagnostic Cost Estimation

Finally, we demonstrate the effectiveness of our cost estimation function. We show that it is often possible to destroy different cones to obtain different abstractions of a system that can all be successfully compiled, and then, using the cost estimation function, select an abstraction to be used for diagnosis that is more likely to give optimal average cost. These results also help explain why in some cases the hierarchical approach causes diagnostic cost to increase compared with the baseline approach.

In these experiments, we use SDA with cloning and include circuits up to $c2670$, considering only single-fault test cases. We did not include the largest circuits in our analysis as these circuits often could not be compiled after some cones in them were destroyed; therefore it was not possible to obtain an overall picture of the actual cost for these circuits. Test cases for circuits up to $c1355$ are the same as used before, whereas for circuits $c1908$ and $c2670$, this time, we use a more complete set of cases as done for smaller circuits. Specifically, we generate $n$ fault scenarios for each circuit, where $n$ equals the number of gates in the circuit: Each scenario contains a different faulty gate. We then randomly generate 1 test case for each of these $n$ scenarios (in some cases, we could not obtain a test case in reasonable time and the corresponding scenarios were not used).

The results of experiments are summarized in Table 7. For each circuit the first row shows results when all cones have been considered and the subsequent rows show results when all cones having more than a specified number of inputs (in column 3) have been destroyed. When the value in column 3 is 0 we get the trivial abstraction, where all cones have been destroyed, which is equivalent to using the baseline approach. The last two columns show the (actual) average cost and time for diagnosing a circuit using the given abstraction. The columns labeled with $AC$, $IC$, and $EDC$ show values obtained using the equations 6, 5, and 7, respectively, for a given abstraction.

The results show that we are often able to destroy several cones while still being able to compile the circuit successfully. However, quite naturally, the compilation time increases as more cones are destroyed such that at some point the circuits start to fail to compile, where we stop destroying cones. The actual diagnostic cost on different circuits show different trends each time some cones have been destroyed. For example, on $c432$ it shows significant improvement while the reverse is true for $c1355$. On remaining circuits the actual cost shows somewhat mixed trends; however, the relative increase or decrease in the costs is generally less significant.

Comparison of the isolation and abstraction costs (i.e., $IC$ and $AC$, respectively) for various abstractions confirms that each time some cones are destroyed the isolation cost increases while the abstraction cost decreases. It is the potentially imbalanced change in the two costs that determines whether the cost might go up or down after the cones are destroyed. For example, in the case of $c432$ the abstraction cost drops more rapidly than the isolation cost increases when cones are destroyed, while in the case of $c1355$ the two costs change almost at the same pace.

Comparison of the predicted costs $EDC$ with the actual costs shows that for $c432$, $c499$, $c1908$, and $c2670$ the predicted costs are often quite close to the actual costs, which demonstrates the relative accuracy of our approach. As a result, for these circuits the cost estimation function can accurately predict the abstraction that is more likely to give optimal cost. For example, it correctly suggests that one should use the baseline approach with $c432$. For the other two circuits, $c880$ and $c1355$, the predicted and actual costs are significantly different, and the cost estimation function fails to give good predictions. $c1355$

seems to be a special case in which the actual diagnostic cost increases quite rapidly as cones are destroyed, the reason for which will be an interesting topic for future work.

## 8. Related Work

Out et al. (1994) considered two kinds of hierarchical models and discussed automatic methods for constructing their abstractions. In the first kind, components of the given detailed model are aggregated into single components of the abstract model, such that every diagnosis of the detailed model, refined from a diagnosis of the abstract model, is guaranteed to be valid. Thus there is no need to check the validity of detailed diagnoses afterwards. In the second kind, the abstract model is constructed such that it is always possible to determine a unique diagnosis at every level of the hierarchy with a reasonable cost, where the measurements that are less costly to make appear in the most abstract model and the more costly measurements appear in the most detailed model. More techniques for automatic abstraction-based on system observability were discussed by Torta and Torasso (2003, 2008). These papers provide alternative techniques to automatic abstraction; however, they do not address sequential diagnosis.

The idea of testing the most likely failing component comes from Heckerman et al. (1995), where the testing of a component was considered a unit operation and components were tested in decreasing order of their likelihood of failure, which was computed assuming a single fault (this assumption could compromise the quality of the measurement sequence in multiple-fault cases as the authors pointed out). In our case, by contrast, the testing of each variable of a component is a unit operation, calling for a more complex heuristic in order to minimize the number of tests; also, we do not need to assume a single fault. Our work also goes further in scalability using several structure-based techniques: compilation, abstraction, and component cloning.

Chittaro & Ranon (2004) considered the computation of diagnoses using a hierarchical algorithm. Their method takes a hierarchical decomposition of the system as input, where sets of components are aggregated into units, and computes a set of diagnoses at the most abstract level, which are then refined hierarchically to the most detailed level. Feldman & van Gemund (2006) developed a hierarchical diagnosis algorithm and tested it on reverse engineered ISCAS-85 circuits (Hansen, Yalcin, & Hayes, 1999) that are available in high-level form. The idea is to decompose the system into hierarchies in such a way as to minimize the sharing of variables between them. This can be done for well engineered problems and they have formed hierarchies by hand for ISCAS-85 circuits. The system is represented by a hierarchical logical formula where each hierarchy is represented by a traditional CNF formula. This representation can be translated to a fully hierarchical DNF, a fully flattened DNF, or a partially flattened DNF dictated by a depth parameter, after which a hierarchical search algorithm is employed to find the diagnoses. The hierarchical aspect of these two approaches is similar to that of ours; however, they require a hierarchical decomposition of the system to be either given as part of the input, or obtained by hand, while our approach searches for hierarchies automatically. Another major difference is that they consider only the computation of diagnoses and do not address the problem of sequential diagnosis.

Based on the GDE framework, de Kleer (2006) studied the sensitivity of diagnostic cost to what is called the $\epsilon$-policy, which is the policy that quantifies how the posterior probabilities of diagnoses are to be estimated when GDE computes its heuristic. In our case, probabilities of diagnoses are not required at all, and the other probabilities that are required can all be computed exactly by evaluating and differentiating the d-DNNF. Nevertheless, our algorithm can be sensitive to the initial probabilistic model given and sensitivity analysis in this regard may lead to interesting findings.

Recently, Flesch, Lucas, & van der Weide (2007) proposed a new framework to integrate probabilistic reasoning into model-based diagnosis. The framework is based upon the notion of *conflict measure*, which originated as a tool for the detection of conflicts between an observation and a given Bayesian network (Jensen, 2001). When a system is modeled as a Bayesian network for diagnostic reasoning, it is possible to use this conflict measure to differentiate between diagnoses according to their degree of consistency with a given set of observations. This work, however, does not address the problem of sequential diagnosis, i.e., locating actual faults by taking measurements.

Most recently, Feldman, Provan, and van Gemund (2009) proposed a related method for reducing diagnostic uncertainty. While our work attempts to identify the actual faults with the fewest individual measurements, their heuristic was aimed at reducing the number of diagnoses with the fewest test vectors.

## 9. Conclusion

We have presented a new system for sequential diagnosis, called SDA, that employs four new structure-based techniques to scale diagnosis to larger systems. Specifically, it uses a heuristic for measurement selection that can be computed efficiently from the d-DNNF compilation of the system. To diagnose larger systems, it automatically computes a structural abstraction of the system and performs diagnosis in a hierarchical fashion. It then employs a structure-based technique for further reducing the abstraction size of the system, which scales the diagnosis to the largest benchmark systems. Finally, it can automatically select an abstraction of the system that is more likely to give optimal average cost.

## Acknowledgments

We thank the anonymous reviewers for their comments. NICTA is funded by the Australian Government as represented by the Department of Broadband, Communications and the Digital Economy and the Australian Research Council through the ICT Centre of Excellence program. Part of this work has appeared in KR 2010 (Siddiqi & Huang, 2010); another part of this work was carried out during July–September 2010 while the first author was visiting NICTA.

## Appendix A. Computing Probabilities on d-DNNF

Here we briefly describe the computation of probabilities based on d-DNNF compilations of Bayesian networks. d-DNNF is a graph representation of a nested and/or expression where negation only appears next to variables, children of every *and*-node have disjoint sets of variables (*decomposability*), and children of every *or*-node are pairwise logically inconsistent

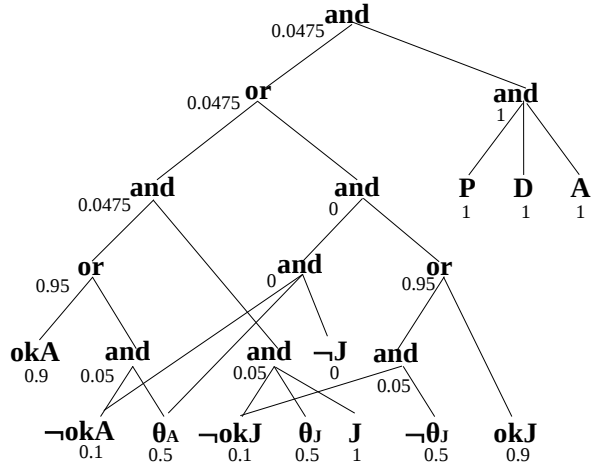

Figure 6: d-DNNF compilation of subcircuit (dotted) in Figure 1 given the observation $A \wedge P \wedge D$ and computation of the posterior probability of $J = 1$.

(*determinism*). For example, Figure 6 shows a d-DNNF compilation of the subcircuit in the dotted box of Figure 1 under the observation $A \wedge P \wedge D$.

Given a d-DNNF compilation, the probability $Pr(\mathbf{E} = \mathbf{e})$ for an instantiation $\mathbf{e}$ of any set of variables $\mathbf{E}$ can be obtained by the following linear-time procedure: (i) Set all variables $\mathbf{E}$ to Boolean constants according to the instantiation $\mathbf{e}$, (ii) set all other literals (not in $\mathbf{E}$) to true except those that have numbers associated with them (negative literals are associated with 1 minus the corresponding numbers for the positive literals), and (iii) evaluate the d-DNNF bottom-up by treating true as 1, false as 0, the remaining leaves as their associated numbers, *or*-nodes as additions, and *and*-nodes as multiplications. The number at the root will be $Pr(\mathbf{E} = \mathbf{e})$. For example, Figure 6 shows the computation of the probability of $J = 1$ given the observation $A \wedge P \wedge D$. Thus $\mathbf{e} = \{A = 1, P = 1, D = 1, J = 1\}$. In the d-DNNF, we set $A = 1, P = 1, D = 1, J = 1, \neg J = 0$. The rest of the literals are given values that are associated with them (discussed in Section 3.2).

Furthermore, a second traversal of the d-DNNF, from the top down, can effectively differentiate the d-DNNF so that updated probabilities are computed at once for every possible change in the value of a variable (e.g., from unknown to known) (Darwiche, 2003). This is useful for our measurement point selection where we need to update the entropies for all candidate measurement points.

## Appendix B. Cardinality-based Model Pruning

Here we present the technique referred to in Section 4 that can be used to remove a significantly large number (if not all) of diagnoses of cardinality $> k$ from the d-DNNF.

The value of $k$ must be greater or equal to the minimum-cardinality of the d-DNNF for pruning to occur. If $k$ is equal to the minimum-cardinality of the d-DNNF then all diagnoses with cardinality $> k$ can be removed using the minimization procedure described by Darwiche (2001). If, however, $k$ is greater than the minimum-cardinality of the d-DNNF then we need a similar but modified minimization algorithm to make sure we do not remove diagnoses of cardinality $\leq k$.

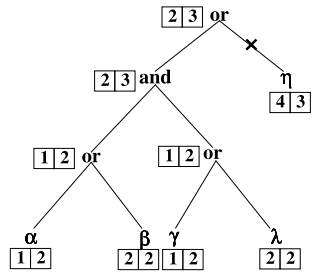

Figure 7: Pruning d-DNNF to improve heuristic accuracy.

While a complete pruning is difficult to achieve in general, an approximation is possible. In a naive approach, one may remove every child $l$ of every $or$-node $n$ for which minimum-cardinality ($mc$) of $l$ is greater than $k$, which will be sound in that it will never remove diagnoses of cardinality $\leq k$ but may result in too little pruning in many cases. We can increase the amount of pruning performed by computing local value $k(n)$ for every node $n$ given the global $k$ for the whole d-DNNF using a top-down traversal through the d-DNNF: Every node $n$ suggests a value $k(l)$ for its child $l$ and the largest of these values is accepted to be the final value of $k(l)$ (this is essential to avoid possibly removing diagnoses of cardinality $\leq k$). More pruning can occur in this way because $k(n)$ can often be less than the global $k$. Once $k(n)$ has been computed for every node, every child $l$ of every $or$-node $n$ for which $mc(l) > k(l)$ can then be pruned.

We now give the pruning algorithm which performs a two pass traversal through the d-DNNF. The $mc(n)$ is updated during upward traversal and represents the minimum-cardinality of diagnoses under a node $n$, whereas the $k(n)$ is updated during downward traversal and represents the upper bound on the fault-cardinality for a node which is used to prune branches emanating from the node whose $mc(n)$ exceeds the $k(n)$.

The two passes of the procedure are as follows: Initialize $mc(n)$ to 0 and $k(n)$ to -$\infty$ (least possible value) for all $n$. Traverse the d-DNNF so that children are visited before parents and for every leaf node, set $mc(n)$ to 1 if $n$ is a negated health variable and 0 otherwise; for every $or$-node, set $mc(n)$ to the minimum of the values of $mc$ of its children; for every $and$-node set $mc(n)$ to the sum of the values of $mc$ of its children. Now traverse the d-DNNF so that parents are visited before children and set $k(n)$ for the root node to the value $k$; for every $or$-node, remove every child $p$ of $n$ for which $mc(p) > k(n)$ and for every remaining child $v$ set $k(v)$ to $k(n)$ if $k(n) > k(v)$; for every child $p$ of every $and$-node, let $t_p$ be the sum of the values of $mc$ of all the other children and set $k(p)$ to the value $t_p$ if $t_p > k(p)$.

In the above procedure the conditions $k(n) > k(v)$ and $t_p > k(p)$ while updating $k$ for a node ensure that only a safe value for $k$ is set. An example is shown in Figure 7. The $mc$ (left) and $k$ (right) values are shown for each node. The branches labeled $\alpha$, $\beta$, $\gamma$, and $\eta$ are subgraphs associated with hypothetical values for $mc$. The figure shows that the minimum-cardinality for every node ($mc$) is less than or equal to the bound ($k$) except for the branch labeled $\eta$, which gets pruned accordingly.

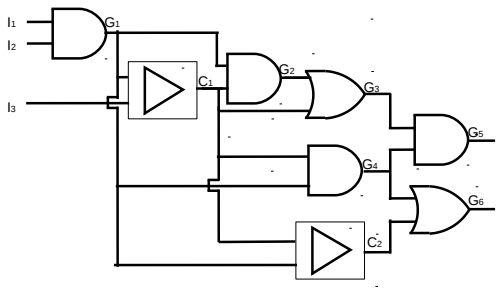

Figure 8: A combinational circuit generated randomly from a set of components consisting of gates $G_1, G_2, \ldots, G_6$ and cones $C_1, C_2$, when they are processed in the order: $G_1, C_1, G_2, G_3, G_4, C_2, G_5, G_6$.

| N | total gates | average depth | approx. treewidth | abstraction size | total clones | abstraction size after cloning |
|---|---|---|---|---|---|---|
| **32** | 104 | 26.9 | 13 | 26 | 32 | 17 |
| **40** | 130 | 31.6 | 16 | 31 | 42 | 20 |
| **48** | 156 | 30.3 | 17 | 38 | 69 | 24 |
| **56** | 182 | 34.8 | 21 | 45 | 68 | 28 |
| **64** | 208 | 37.6 | 24 | 51 | 84 | 32 |
| **72** | 234 | 41.1 | 26 | 59 | 108 | 38 |
| **80** | 260 | 39.3 | 29 | 66 | 128 | 41 |
| **88** | 286 | 41.6 | 32 | 71 | 158 | 42 |
| **96** | 312 | 46.3 | 34 | 79 | 172 | 48 |
| **104** | 338 | 43.4 | 36 | 82 | 177 | 49 |
| **112** | 364 | 41.8 | 39 | 90 | 194 | 57 |
| **120** | 390 | 48.5 | 43 | 97 | 218 | 61 |
| **128** | 416 | 48.1 | 45 | 104 | 194 | 65 |
| **136** | 442 | 50.7 | 48 | 112 | 243 | 72 |
| **144** | 468 | 48.2 | 51 | 116 | 265 | 70 |
| **152** | 494 | 50.8 | 51 | 123 | 272 | 78 |

Table 8: Randomly generated combinational circuits $(N, 25, 5)$.

## Appendix C. Randomly Generated Combinational Circuits

In this section we use a novel method to systematically generate series of combinational circuits such that their structure and size can be controlled. This enables the evaluation of our techniques on circuits other than ISCAS-85 benchmarks, which has helped us identify factors that affect the diagnostic cost, leading us to the cost estimation function given in Section 6. Specifically, we observe that for circuits of a similar structure, diagnostic cost generally increases with circuit size, which helped us devise the notion of isolation cost; and that when circuit size is held constant, diagnostic cost generally increases with the number of cones in the circuit, which helped us devise the notion of abstraction cost.

The circuits are generated by composing a set of pre-formed building blocks. The latter consist of both gates and cones. The gates are taken from a pool of six gates of types OR, NOR, AND, NAND, NOT, and BUFFER, and the cones from a pool of eight cones, each of which has 10 gates and is extracted from ISCAS-85 benchmark circuits.

Our composition method is inspired from the method of generating random Bayesian networks described by Marinescu, Kask, and Dechter (2003). The circuits are generated according to a formula $(N, P, I)$, where $N$ is the number of components (building blocks) to use, $P$ the percentage of cones in the components, and $I$ the maximum number of inputs a gate can have. To generate the $N$ components we randomly pick $(P/100) * N$ cones (with repetition) from the pool of cones and $N - (P/100) * N$ gates (with repetition) from the pool of gates and place them in a random order. The number of inputs of each gate is set randomly between 2 and $I$, except for a NOT or BUFFER gate which can have only one input.

We then process each component as follows: Suppose that the components are placed in the order $C_1, C_2, \ldots, C_N$. Let $\mathbf{P_i}$ be the set of components that precede $C_i$ in the order. When we process a component $C_i$ we connect every input of $C_i$ to the output of a randomly chosen component from $\mathbf{P_i}$ such that no two inputs of $C_i$ are connected to the same component. If an input of $C_i$ cannot be connected (either because $\mathbf{P_i}$ is empty or all the components in $\mathbf{P_i}$ have been used) then it is treated as a primary input of the circuit. For example, the circuit in Figure 8 has been randomly generated according to the formula $(8, 25, 2)$, where the components shown in the boxes represent cones.

By varying the parameters $(N, P, I)$ we can obtain circuits of varying size and structure. First we fix $P = 25$, $I = 5$ and vary $N$ to generate a range of circuits of increasing size. For each $N$ we generate 10 circuits. These circuits are summarized in Table 8. The numbers in the columns are averaged over all circuits of a given size, and rounded off. Generally, when $N$ is increased we see an increase in the abstraction size as well as the estimated treewidth, corresponding to an increase in the perceived difficulty of the circuit (e.g., note that the largest circuit in this set is smaller than $c1355$, but the estimated treewidth of $c1355$ is much lower, at 25; the actual compilation was indeed harder for the former circuit). For each circuit we randomly generate 10 single-fault, 10 double-fault, and 10 five-fault scenarios and a single test case for each scenario.

The results of experiments with these circuits are given in Tables 9, 10, and 11, using the baseline, hierarchical, and cloning techniques, respectively. These results are generally consistent with those obtained using the ISCAS-85 circuits. The baseline SDA could not solve any circuit beyond $(72, 25, 5)$. The hierarchical SDA solved more circuits but could not solve any circuit beyond $(80, 25, 5)$. The most scalable version of SDA, with component cloning, solved much larger circuits, up to $(168, 25, 5)$.

Note that there is a general trend of increase in diagnostic cost with increase in $N$. This is consistent with one's intuitive expectation that diagnostic uncertainty would increase with system size. Also note that diagnostic cost is often significantly higher for the hierarchical approach than the baseline approach. As discussed earlier, this can be attributed to the fact that the hierarchical approach often has to go through hierarchies of cones to reach a faulty gate, which the baseline approach may be able to reach more directly.

| N | total gates | pruning | single-fault | | | double-fault | | | five-fault | | |
|---|---|---|---|---|---|---|---|---|---|---|---|
| | | | *solved* | *cost* | *time* | *solved* | *cost* | *time* | *solved* | *cost* | *time* |
| **32** | 104 | *no* | 100 | 5.86 | 0.56 | 100 | 6.34 | 0.57 | 100 | 9.19 | 0.60 |
| | | *yes* | 100 | 4.81 | 0.54 | 100 | 5.24 | 0.55 | 100 | 8.22 | 0.59 |
| **40** | 130 | *no* | 100 | 5.82 | 4.31 | 100 | 7.05 | 4.51 | 100 | 11.53 | 5.09 |
| | | *yes* | 100 | 4.5 | 4.16 | 100 | 5.08 | 4.28 | 100 | 10.35 | 4.93 |
| **48** | 156 | *no* | 100 | 6.58 | 32.43 | 100 | 8.72 | 32.75 | 100 | 11.19 | 34.87 |
| | | *yes* | 100 | 4.73 | 31.27 | 100 | 5.9 | 31.14 | 100 | 9.46 | 33.84 |
| **56** | 182 | *no* | 80 | 5.26 | 190.99 | 80 | 6.9 | 192.69 | 80 | 11.05 | 202.4 |
| | | *yes* | 80 | 3.58 | 185.32 | 80 | 5.62 | 190.25 | 80 | 8.325 | 197.05 |
| **64** | 208 | *no* | 50 | 5.58 | 532.82 | 50 | 6.9 | 540.31 | 50 | 13.94 | 581.11 |
| | | *yes* | 50 | 5.02 | 527.24 | 50 | 4.72 | 525.02 | 50 | 9.84 | 558.79 |
| **72** | 234 | *no* | 10 | 6.2 | 207.89 | 10 | 9.5 | 230.72 | 10 | 27.5 | 354.80 |
| | | *yes* | 10 | 6.2 | 207.49 | 10 | 5.8 | 205.41 | 10 | 11.4 | 248.20 |

Table 9: Baseline heuristic on randomly generated circuits $(N, 25, 5)$.

| N | total gates | pruning | single-fault | | | double-fault | | | five-fault | | |
|---|---|---|---|---|---|---|---|---|---|---|---|
| | | | *solved* | *cost* | *time* | *solved* | *cost* | *time* | *solved* | *cost* | *time* |
| **32** | 104 | *no* | 100 | 7.81 | 0.15 | 100 | 8.78 | 0.16 | 100 | 12.59 | 0.18 |
| | | *yes* | 100 | 3.42 | 0.15 | 100 | 5.87 | 0.16 | 100 | 11.88 | 0.17 |
| **40** | 130 | *no* | 100 | 7.2 | 0.71 | 100 | 8.19 | 0.72 | 100 | 13.77 | 0.75 |
| | | *yes* | 100 | 3.07 | 0.70 | 100 | 5.18 | 0.71 | 100 | 12.94 | 0.73 |
| **48** | 156 | *no* | 100 | 7.03 | 4.10 | 100 | 8.12 | 4.14 | 100 | 12.78 | 4.26 |
| | | *yes* | 100 | 3.18 | 4.01 | 100 | 4.96 | 4.02 | 100 | 11.51 | 4.08 |
| **56** | 182 | *no* | 100 | 7.81 | 42.63 | 100 | 9.1 | 43.58 | 100 | 11.92 | 43.64 |
| | | *yes* | 100 | 2.98 | 41.60 | 100 | 6.31 | 42.23 | 100 | 11.1 | 42.19 |
| **64** | 208 | *no* | 80 | 8.35 | 108.61 | 80 | 9.11 | 107.96 | 80 | 14.85 | 111.04 |
| | | *yes* | 80 | 3.31 | 107.05 | 80 | 5.35 | 106.31 | 80 | 13.56 | 107.71 |
| **72** | 234 | *no* | 30 | 7.56 | 120.59 | 30 | 9.83 | 122.50 | 30 | 12.66 | 123.81 |
| | | *yes* | 30 | 2.8 | 118.35 | 30 | 5.53 | 118.57 | 30 | 11.2 | 119.93 |
| **80** | 260 | *no* | 10 | 6.9 | 190.66 | 10 | 9.2 | 193.58 | 10 | 12.4 | 197.29 |
| | | *yes* | 10 | 2.8 | 188.95 | 10 | 4.6 | 189.73 | 10 | 10.5 | 190.07 |

Table 10: Hierarchical heuristic on randomly generated circuits $(N, 25, 5)$.

We also observe that, again, pruning leads to a general improvement in diagnostic cost. The improvement is more significant for the hierarchical approach, which can be explained by the fact that the effect of pruning is much greater on the abstract model, as each branch pruned can correspond to a large part of the original system.

We now perform another set of experiments to study the impact of hierarchy in a controlled manner. This time we hold the size of the circuits more or less constant and vary the percentage of cones in them. Specifically, we generate a large number of random circuits with $P$ ranging from 0 to 50, such that for each value of $P$ the generated circuits contain 120 gates on average.

The experiments on these circuits are summarized in Table 12. Note that as $P$ increases the estimated treewidth of the circuits decreases, as would be expected, and the actual compilation time indeed also decreases. The diagnostic cost, on the other hand, increases steadily up to $P = 25$ and remains more or less flat afterwards. This confirms the potential negative impact of hierarchy on the diagnostic cost we hypothesized: As $P$ increases the likelihood of a fault occurring inside a cone also increases and thus on average one has to take more measurements, many on inputs to cones, to locate a fault. That diagnostic cost does not further increase after $P = 25$ is consistent with the observation that since the circuit size is fixed at roughly 120 and each cone contributes 10 gates to the circuit, when $P$ increases to some point, there will be very few gates lying outside cones and hence the likelihood of a fault occurring in a cone will have more or less plateaued.

| N | total gates | single-fault | | | double-fault | | | five-fault | | |
|---|---|---|---|---|---|---|---|---|---|---|
| | | *solved* | *cost* | *time* | *solved* | *cost* | *time* | *solved* | *cost* | *time* |
| **32** | 104 | 100 | 7.86 | 0.04 | 100 | 8.78 | 0.05 | 100 | 12.13 | 0.06 |
| **40** | 130 | 100 | 8.12 | 0.05 | 100 | 9.6 | 0.06 | 100 | 13.58 | 0.08 |
| **48** | 156 | 100 | 8.25 | 0.07 | 100 | 9.34 | 0.08 | 100 | 12.6 | 0.10 |
| **56** | 182 | 100 | 9.03 | 0.12 | 100 | 10.4 | 0.13 | 100 | 13.37 | 0.15 |
| **64** | 208 | 100 | 10.06 | 0.45 | 100 | 10.73 | 0.46 | 100 | 15.41 | 0.49 |
| **72** | 234 | 100 | 9.15 | 0.78 | 100 | 11.38 | 0.80 | 100 | 15.44 | 0.84 |
| **80** | 260 | 100 | 9.78 | 0.83 | 100 | 11.38 | 0.85 | 100 | 15.5 | 0.89 |
| **88** | 286 | 100 | 9.56 | 0.78 | 100 | 10.87 | 0.79 | 100 | 16.6 | 0.84 |
| **96** | 312 | 100 | 10.4 | 1.85 | 100 | 10.81 | 1.87 | 100 | 17.87 | 1.97 |
| **104** | 338 | 100 | 10.03 | 4.23 | 100 | 11.79 | 4.26 | 100 | 16.95 | 4.34 |
| **112** | 364 | 100 | 10.44 | 29.20 | 100 | 11.76 | 29.39 | 100 | 17.62 | 29.93 |
| **120** | 390 | 100 | 10.36 | 39.88 | 100 | 13.6 | 40.15 | 100 | 20.76 | 41.17 |
| **128** | 416 | 90 | 11.17 | 98.70 | 90 | 13.73 | 99.08 | 90 | 19.33 | 100.73 |
| **136** | 442 | 90 | 11.82 | 220.41 | 90 | 13.76 | 221.63 | 89 | 20.25 | 225.58 |
| **144** | 468 | 80 | 12.08 | 207.69 | 80 | 15.05 | 207.68 | 80 | 19.92 | 210.86 |
| **152** | 494 | 40 | 12.7 | 256.43 | 40 | 14.72 | 257.5 | 40 | 23.02 | 260.41 |
| **160** | 520 | 40 | 12.5 | 476.93 | 40 | 14.15 | 479.33 | 40 | 18.5 | 479.83 |
| **168** | 546 | 10 | 8.7 | 84.16 | 10 | 10.1 | 84.44 | 10 | 15.1 | 85.27 |

Table 11: Component cloning on randomly generated circuits ($N$, 25, 5).

| P | total circuits | treewidth | single-fault | | double-fault | | five-fault | |
|---|---|---|---|---|---|---|---|---|
| | | | *cost* | *time* | *cost* | *time* | *cost* | *time* |
| 0 | 1000 | 32 | 5.7 | 8.8 | 7.7 | 8.8 | 13.5 | 9.0 |
| 5 | 600 | 23 | 6.7 | 0.9 | 8.0 | 0.9 | 13.0 | 0.9 |
| 10 | 900 | 21 | 7.5 | 0.5 | 8.7 | 0.5 | 13.2 | 0.5 |
| 15 | 1000 | 18 | 7.7 | 0.1 | 9.1 | 0.1 | 12.2 | 0.1 |
| 20 | 1100 | 17 | 8.0 | 0.1 | 9.4 | 0.1 | 12.3 | 0.1 |
| 25 | 1300 | 15 | 9.2 | 0.1 | 10.3 | 0.1 | 13.6 | 0.1 |
| 50 | 800 | 12 | 8.6 | 0.06 | 10.0 | 0.07 | 12.4 | 0.1 |

Table 12: Component cloning on randomly generated circuits ($N$,P,5).